\documentclass[journal]{IEEEtran}
\usepackage{amsmath,amsfonts}
\usepackage{algorithmic}
\usepackage{array}
\usepackage{booktabs}
\usepackage{multirow}
\usepackage[caption=false,font=normalsize,labelfont=sf,textfont=sf]{subfig}
\usepackage{textcomp}
\usepackage{stfloats}
\usepackage{url}
\usepackage{verbatim}
\usepackage{graphicx}
\usepackage{cite}

\usepackage{color}
\usepackage{xspace}
\newcommand*{\proposed}{FreqSal\xspace}

\usepackage{hyperref}
\hypersetup{hidelinks, colorlinks=True, allcolors=magenta, pdfstartview=Fit, breaklinks=true}

\def\BibTeX{{\rm B\kern-.05em{\sc i\kern-.025em b}\kern-.08em
    T\kern-.1667em\lower.7ex\hbox{E}\kern-.125emX}}
\usepackage{balance}

\begin{document}
\title{Deep Fourier-embedded Network for RGB and Thermal Salient Object Detection}

\author{Pengfei Lyu, Xiaosheng Yu, Pak-Hei Yeung, Chengdong Wu, and Jagath C. Rajapakse, \IEEEmembership{Fellow, IEEE}

\thanks{This work was supported in part by the National Natural Science Foundation of China under Grant U20A20197 and 62306187, the Foundation of Ministry of Industry and Information Technology TC220H05X-04, Liaoning Key Research and Development Project under Grant 2020JH2/10100040, Natural Science Foundation of Liaoning Province under Grant 2021-KF-12-01, the Foundation of National Key Laboratory under Grant OEIP-O-202005, the Fundamental Research Fund for the Central Universities of China under Grant N2326001 and the China Scholarship Fund. \emph{(Corresponding author: Chengdong Wu and Jagath C. Rajapakse).}

Pengfei Lyu is with the Faculty of Robot Science and Engineering, Northeastern University, Shenyang, 110169 China, and also with the College of Computing and Data Science, Nanyang Technological University, 639798 Singapore (e-mail:lyupengfei@stumail.neu.edu.cn).

Xiaosheng Yu and Chengdong Wu are with the Faculty of Robot Science and Engineering, Northeastern University, Shenyang, 110169 China (e-mail: yuxiaosheng@mail.neu.edu.cn; wuchengdong@mail.neu.edu.cn).

Pak-Hei Yeung and Jagath C. Rajapakse are with the College of Computing and Data Science, Nanyang Technological University, 639798 Singapore (e-mail: pakhei.yeung@ntu.edu.sg; asjagath@ntu.edu.sg).}}

\markboth{Journal of \LaTeX\ Class Files,~Vol.~18, No.~9, September~2020}%
{How to Use the IEEEtran \LaTeX \ Templates}
\maketitle

\begin{abstract}
The rapid development of deep learning has significantly improved salient object detection (SOD) combining both RGB and thermal (RGB-T) images. 
However, 
existing Transformer-based RGB-T SOD models with quadratic complexity are 
memory-intensive,
limiting their application in high-resolution bimodal feature fusion. 
To overcome 
this limitation,
we propose a purely Fourier Transform-based model, namely Deep Fourier-embedded Network (\proposed), for accurate RGB-T SOD. 
Specifically, we leverage the efficiency of Fast Fourier Transform with linear complexity to design three key components: 
(1) To fuse RGB and thermal modalities, we propose Modal-coordinated Perception Attention, which aligns and enhances bimodal Fourier representation in multiple dimensions; (2) To clarify object edges and suppress noise, we design Frequency-decomposed Edge-aware Block, which deeply decomposes and filters Fourier components of low-level features; (3) To accurately decode features, we propose Fourier Residual Channel Attention Block, which prioritizes high-frequency information while aligning channel-wise global relationships.
Additionally, even when converged, existing deep learning-based SOD models' predictions still exhibit frequency gaps relative to ground-truth.
To address this problem,
we propose Co-focus Frequency Loss, which dynamically weights hard frequencies during edge frequency reconstruction by cross-referencing bimodal edge information in the Fourier domain. 
Extensive experiments on ten bimodal SOD benchmark datasets demonstrate that \proposed outperforms twenty-nine existing state-of-the-art bimodal SOD models. Comprehensive ablation studies further validate the value and effectiveness of our newly proposed components. The code is available at \url{https://github.com/JoshuaLPF/FreqSal}.
\end{abstract}

\section{Introduction}
\IEEEPARstart{S}{alient} object detection (SOD) identifies the most attention-grabbing objects in a scene, with broad applications spanning image compression \cite{5223506}, object recognition \cite{flores2019saliency}, scene perception \cite{10182276}, and more. 
Despite significant progress being made in RGB-based SOD  
\cite{zhou2023texture,wang2023pixels}, 
it remains challenging to separate salient objects from complex backgrounds or low-contrast scenes.
To address this challenge, depth information was introduced to complement RGB features, enhancing the understanding of scenes 
\cite{VST,NLPR,NJUD,DUT-RGBD,SIP,STERE}. 

However, the accuracy and stability of depth information are susceptible to environmental factors, such as varying light intensity, occlusions, and other conditions, limiting its applicability. 
In contrast, thermal infrared information offers distinct advantages in environmental adaptability. 
Even under adverse weather or low-light conditions, 
thermal images can effectively characterize objects by capturing their infrared radiation. As a result, RGB and thermal (RGB-T) SOD has gained increasing attention \cite{wang2018rgb,8744296,9767629,10003255,EDEF}. 

The Transformer architecture\cite{vaswani2017attention} has demonstrated excellent ability to acquire global dependencies, outperforming convolutional neural networks (CNNs), which are constrained by local receptive fields,
in various computer vision tasks \cite{dosovitskiy2021an,liu2021swin}. 
This success has led to the development of Transformer-based models for RGB-T SOD, which have achieved impressive performance \cite{10015881,9611276,9869666,chen2022cgmdrnet,10015667,zhou2023position}. 
However, a significant challenge arises from the high memory consumption 
of Transformers, 
particularly when handling high-resolution bimodal features 
\cite{zhou2023position,TwinsTNet}. 
To mitigate the memory footprint 
for training with limited computational resources,
existing Transformer-based accuracy-targeted models\footnote{Accuracy-targeted models focus on achieving high performance and are typically heavyweight, whereas lightweight models emphasize efficiency.} usually employ compromise strategies, such as 
sacrificing the exploration of relationships between higher resolution bimodal features \cite{VST,zhou2023position,liu2024vst++,UniTR,SACNet},
constructing CNN-based fusion modules \cite{10015881,9611276,chen2022cgmdrnet,9869666,jin2024cafcnet}, 
and utilizing CNN-based backbones \cite{chen2022cgmdrnet,10015667,jin2024cafcnet}\footnote{The simplistic structure diagrams of existing Transformer-based accuracy-targeted models are presented in the Appendix.}. 
These compromise strategies result in models only partially explore global relationships, leading to less effective performance. 
This raises the question: 
Is there a more efficient operation 
that captures global relationships across each part of the model while avoiding excessive memory consumption?

Recently, benefiting from its powerful global representation capabilities, Fast Fourier transform (FFT) \cite{frigo1998fftw} has shown remarkable performance in image enhancement tasks \cite{huang2022deep,li2023embedding,wang2023fourllie, huang2024revitalizing}. 
Importantly, FFT is a global-based transform that represents all previous elements with each transformed element, making it memory-efficient.
\begin{figure}[!htp]
	\centering \includegraphics[width=0.48\textwidth]{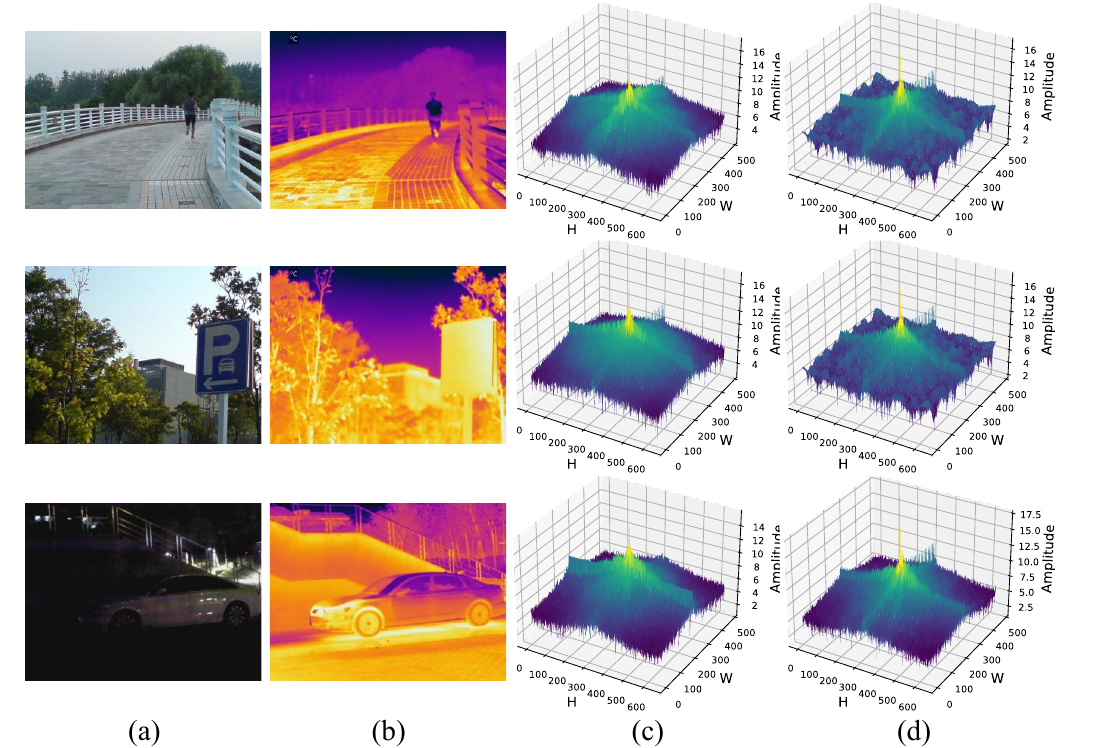}
	\caption{Frequency spectrum visualization: (a) RGB and (b) thermal images, with (c) and (d) showing their corresponding frequency spectrograms.}
\label{fig:spectrum}
\end{figure}
Inspired by this observation, we propose the purely FFT-based architecture, \textbf{\proposed}, 
aiming to align global dependencies at each stage---including encoding, modal fusion, decoding, and edge detection---while achieving high-resolution bimodal feature fusion and accurate RGB-T SOD.

The differences in imaging techniques between RGB and thermal modalities result in vast disparities in information distribution. 
For example, the frequency spectrograms of RGB and thermal images in Fig. \ref{fig:spectrum} reveal notable differences, reflecting their distinct frequency distributions of saliency.
Specifically, while both modalities depict the same salient objects, they differ in the degree of saliency in the spatial dimension and exhibit distinct patterns in the channel dimension \cite{9611276,CCAFNet}. 
To address this problem, we propose \textbf{Modal-coordinated Perception Attention (MPA)}, a Transformer-like module that aligns and fuses the two modalities across both spatial and channel dimensions in the frequency domain through a re-embedding strategy. 

Given that SOD is a pixel-level dense prediction task, the decoding stage needs to integrate low-resolution encoder features from different levels to predict the high-resolution saliency map with detailed information. 
Therefore, it is important to prioritize high-frequency information in the decoding stage. 
However, the low-resolution high-level features tend to favor semantic information, 
while low-level features possess rich edge details but are accompanied by noisy textures.
To overcome these challenges, 
we propose two novel components: the \textbf{Frequency-decomposed Edge-aware Block (FEB)} and the \textbf{Fourier Residual Channel Attention Block (FRCAB)}.
Specifically, the FEB is designed
to clarify edge features amidst cluttered textures by deeply decomposing low-level features, thereby guiding the integration of decoder features.
Meanwhile, the FRCAB is applied at each decoder layer to mitigate excessive low-frequency pattern information, allowing the model to focus on learning high-frequency details. 
By considering global dependencies between channels in the frequency domain, FRCAB adaptively adjusts the channel features, enhancing the model's discriminative learning capability.

Furthermore, 
existing deep learning-based SOD models \cite{liu2024vst++,UniTR,SACNet,TwinsTNet,SENet,BCARNet}
often lose vital visual details as the network deepens. 
In addition, reconstructing high-frequency edge information is prone to spectral bias \cite{rahaman2019spectral},
causing these models to 
favor learning easier frequencies over the hard ones. 
However, existing edge modules based on spatial domain learning tend to generate sparse spectra \cite{9611276,10042233,10127616},
treating all frequencies equally and 
making it difficult to accurately characterize edges. 
To address these issues, inspired by \cite{lin2017focal,jiang2021focal,hu2022hdnet}, 
we propose the \textbf{Co-focus Frequency Loss (CFL)} to guide FEB in the frequency domain while optimizing the entire model. 
In CFL, we measure the prediction and ground-truth using the Fourier frequency distance and introduce original edge frequency information of RGB and thermal modalities 
as a cross-reference to dynamically weight hard frequencies.
This frequency domain learning approach leads to refined edge features that further improve the quality of the saliency map in spatial domain learning.

By integrating these components, \proposed efficiently fuses RGB and thermal modalities by utilizing complementary relationships from the frequency perspective, ultimately producing saliency maps with high accuracy. In summary, the main contributions of this paper are as follows:   
\begin{itemize}
    \item We propose \proposed, an FFT-based model designed to achieve high-resolution bimodal feature fusion and RGB-T SOD while minimizing memory consumption. Through comprehensive ablation studies, we verify the effectiveness of the proposed components, including MPA, FEB, FRCAB, and CFL. 
    \item We conduct extensive experiments on the ten benchmark datasets and show that our proposed \proposed achieves superior performance and robustness over twenty-nine existing state-of-the-art models. Notably, \proposed is able to process bimodal inputs with resolution up to $512 \times 512$ on an NVIDIA Tesla P100 GPU with 16GB of memory.
    \item To the best of our knowledge, \proposed is the first FFT-based supervised model for SOD tasks. This work demonstrates the potential of FFT-based operations for dense prediction, highlighting their unique advantages while also exploring their limited compatibility with spatial domain operations.
\end{itemize}

\section{Related Work}
\subsection{RGB and Depth Salient Object Detection (RGB-D SOD)}
\label{sec:related_rgbd}
To address the limitations of RGB-based salient object detection, depth information is introduced to better perceive the shape and spatial relationships of objects. Existing RGB-D SOD models can be broadly categorized into two types: CNN-based and Transformer-based. 

CNN-based RGB-D SOD models typically leverage foundational attention mechanisms, 
such as convolutional block attention module \cite{woo2018cbam}  utilized in 
\cite{li2021hierarchical,HiDANet},
dilated convolution \cite{chen2017deeplab} employed in \cite{pang2020hierarchical,cheng2022depth} and atrous spatial pyramid pooling \cite{Yu_2017_CVPR} utilized in \cite{zhang2021rgb,liu2021learning}. 
In addition, Ji et al. \cite{DUT-RGBD} utilized depth information to guide recurrent attention networks.
Furthermore, a three-stream learning network was proposed in \cite{SIP} to automatically filter out low-quality depth maps.

To address the limitations imposed by the local connectivity of CNNs, several Transformer-based models \cite{vaswani2017attention,liu2021swin} have been proposed. Liu et al. \cite{liu2021tritransnet} introduced three Vision Transformers \cite{dosovitskiy2021an} with shared weights to refine high-level fused features. 
The pure Transformer networks were proposed in \cite{VST,liu2024vst++} for simultaneous salient object and edge detection. 
Several works \cite{sun2023catnet,HFMDNet,PICRNet} employed the Swin Transformer \cite{liu2021swin} as a dual-stream encoder to obtain compact feature representations.
Jin et al. \cite{SPDE} proposed a two-stage ``simple-to-complex'' strategy for learning RGB-D samples.
In contrast to these Transformer-based spatial domain models, our proposed \proposed employed 
FFT-based operations to solve bimodal SOD in the frequency domain.

\subsection{RGB and Thermal Salient Object Detection (RGB-T SOD)}
In complex environments with varying lighting conditions or occlusions, 
depth information become unreliable. 
To address this limitation, researchers \cite{wang2018rgb,8744296,9767629} have introduced thermal modality as an alternative to depth information for SOD.
Thermal imaging captures the infrared radiation emitted by objects and is more adaptable to different environmental conditions. 

Similar to RGB-D SOD (Sec. \ref{sec:related_rgbd}), recent efforts in RGB-T SOD primarily focus on inter-modal fusion. 
Several CNN-based fusion strategies have been proposed 
\cite{wang2021cgfnet,9454273,10003255,chen2022modality,wang2024learning,jin2024cafcnet} 
to combine multi-modal, multi-scale, and multi-granular features. 
In addition, Tu et al. \cite{tu2022weakly} proposed a weak alignment model for RGB and thermal modalities. 
Cong et al. \cite{9926193} re-examined the role of thermal modality in relation to light variations.

To capture long-range relationships, researchers have introduced Transformers for various purposes, including use as backbones \cite{10015881,9611276,9869666,TwinsTNet,TCINet,PATNet}, post-processing of fused features \cite{chen2022cgmdrnet,9869666}, or feature fusion \cite{10015667}. 
In \cite{ConTriNet,ISMNet}, triple-flow networks were proposed to mitigate the effects of modal bias and inherent noise.
Wang et al. \cite{SACNet} proposed a semantics-guided attention to identify the relevance of unaligned modalities.
However, the substantial memory consumption of Multi-head Self-attention (MHSA) makes it challenging to apply Transformers in each stage of the model. 
A pure transformer model \cite{zhou2023position} using the Swin Transformer \cite{liu2021swin} in both encoder and decoder 
required four RTX3090 GPUs for training.

To address the expensive computation, several studies \cite{9505635,9803225,10042233,AlignSal,MAGNet} have explored lightweight modeling, but their performance has often fallen short of expectations. 
Zhou et al. \cite{10127616} designed a wavelet transform-based model to acquire saliency knowledge from a teacher model \cite{9611276}. 
Despite its merits, the wavelet transform is inherently locally based and fails to capture global dependencies effectively.

Considering these limitations, we design a purely FFT-based \proposed. 
The FFT is a global transform that has an advantage over MHSA and the wavelet transform in efficiently obtaining global relationships. 
This property allows each component of \proposed to access and optimize global relationships effectively. 
Moreover, \proposed offers memory efficiency and supports input resolution up to $512^2$ on an NVIDIA
Tesla P100 GPU with 16GB of memory (see Sec. \ref{sec:Ablation Study} for details).

\subsection{Fast Fourier Transform (FFT)-based Networks}
Despite the recent advancement achieved by MHSA-based Transformers \cite{dosovitskiy2021an, liu2021swin} in various tasks, 
the trade-off between the perceived field size and computational cost remains a challenge. To address this problem, 
researchers have explored alternative approaches based on the FFT.
For example, Rao et al. \cite{rao2021global} proposed an FFT-based global filter to learn long-distance dependencies with linear complexity. 
Guibas et al. \cite{guibas2021adaptive} constructed a token mixer as a globally continuous convolution,
while Tatsunami et al. \cite{tatsunami2024fft} incorporated the concept of data correlation into the filter to dynamically extract image features. 
In parallel, researchers have also applied FFT-based models for various downstream tasks, including image enhancement \cite{huang2022deep,li2023embedding, wang2023fourllie, huang2024revitalizing}, image inpainting \cite{chu2023rethinking}, image compression \cite{hu2022hdnet}, and image reconstruction \cite{wu2023neural, wang2023spatial}. 
However, there is a notable scarcity of research on dense prediction tasks.
Existing dense prediction models \cite{farshad2022net, 10177799} typically focus on high- and low-frequency components of the Fourier signal, 
neglecting deeper aspects such as amplitude and phase. 

In this paper, we proposed \proposed to develop FFT for addressing dense prediction tasks, specifically SOD.
To the best of our knowledge, this is the first FFT-based supervised model specifically designed for this task.

\begin{figure*}[!htp]
	\centering \includegraphics[width=0.9\textwidth]{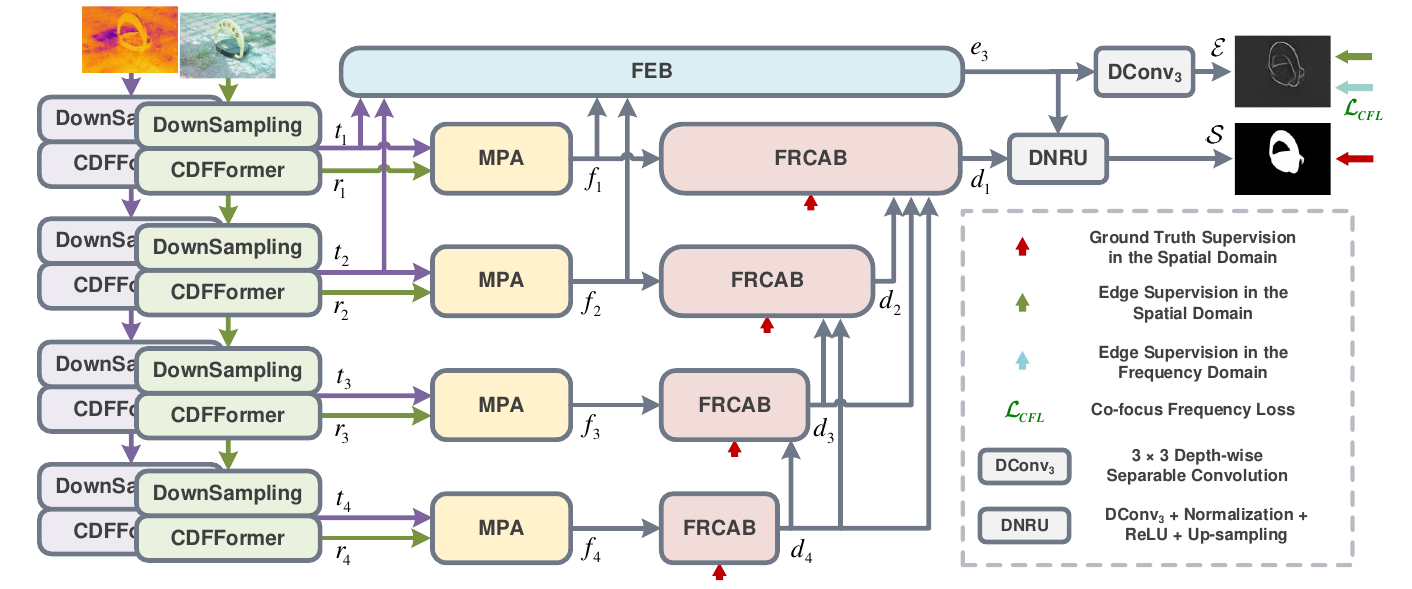}
	\caption{The framework of our proposed \proposed. RGB and thermal images are first extracted initial features $\left\{ {{r_i}} \right\}_{i = 1}^4$ and $\left\{ {{t_i}} \right\}_{i = 1}^4$ by a dual-stream encoder. Next, the bimodal features enter the Modal-coordinated Perception Attention (MPA) to bridge the complementary information and fuse. During decoding, the fused features $\left\{ {{f_i}} \right\}_{i = 1}^4$ are integrated in a low- to high-resolution manner by Fourier Residual Channel Attention Block (FRCAB) to obtain decoder features $\left\{ {{d_i}} \right\}_{i = 1}^4$. Meanwhile, the shallow features $\left\{ {{t_i, f_i}} \right\}_{i = 1}^2$ are fed into the Frequency-decomposed Edge-aware Block (FEB) to obtain elaborate edge features $\left\{ {{e_i}} \right\}_{i = 1}^3$ for guiding the decoding process in generating the accurate saliency map $\mathcal{S}$. In the training stage, the Co-focus Frequency Loss (CFL) ${{\cal L}_{CFL}}$ favoring the synthesis of difficult frequencies and spatial-domain losses are employed together to supervise the generation of the high-quality edge map $\mathcal{E}$.}
\label{fig:frame}
\end{figure*}

\section{Methods}
\subsection{Overview}
In this section, we introduce the proposed \proposed, a novel architecture designed for 
RGB-T SOD, as illustrated in Fig. \ref{fig:frame}. 
The network begins with a dual-stream encoder based on 
FFT-based CDFFormer-M \cite{tatsunami2024fft}, which processes the RGB and thermal modalities separately to extract a set of features at four different scales, 
denoted as $\left\{ {{r_i}} \right\}_{i = 1}^4$ for RGB features and $\left\{ {{t_i}} \right\}_{i = 1}^4$ for thermal features.
The extracted features are then processed by our 
MPA (Sec. \ref{sec:MPA}), which is designed to obtain cross-modal fusion features $\left\{ {{f_i}} \right\}_{i = 1}^4$.  
Subsequently, $\left\{ {{t_i},{f_i}} \right\}_{i = 1}^2$ are input into our 
FEB (Sec. \ref{sec:FEB}), which is designed to extract detailed edge features $\left\{ {{e_i}} \right\}_{i = 1}^3$.
In the decoding stage, 
we employ a pyramid design to handle object scale variations.
Each layer of the decoder concatenates the corresponding fusion feature ${f_i}$ with features from all previous layers. These concatenated features are then refined using our FRCAB (Sec. \ref{sec:FRCAB}), resulting in refined features $\left\{ {{d_i}} \right\}_{i = 1}^4$. 
Finally, the network is optimized using a bi-domain learning approach (Sec. \ref{sec:bidomain}), which combines the proposed 
CFL ${{\cal L}_{CFL}}$, binary cross-entropy loss (BCE) \cite{de2005tutorial} ${{\cal L}_{BCE}}$, and intersection-over union loss  (IoU) \cite{mattyus2017deeproadmapper} ${{\cal L}_{IoU}}$. 
${{\cal L}_{CFL}}$ in the frequency domain facilitates global optimization, prioritizing the resolution of hard frequencies during the edge-frequency reconstruction process. 
Additionally, ${{\cal L}_{BCE}}$ and ${{\cal L}_{IoU}}$ in the spatial domain achieve substantial pixel-wise saliency prediction $\mathcal{S}$.

\subsection{Modal-coordinated Perception Attention (MPA)}
\label{sec:MPA}
The differences in saliency distribution between RGB and thermal modalities in the frequency domain stem from their distinct physical properties, encapsulating complementary information. 
Specifically, while objects share the same spatial location across both modalities, they exhibit varying degrees of saliency, leading to differences in spatial representation.
The RGB modality provides richer appearance and texture details whereas the thermal modality emphasizes overall shape and structure. 
As a result, the importance of feature channels differs across modalities. This observation highlights the need to align and integrate bimodal features in both spatial and channel dimensions \cite{9611276,CCAFNet}. 

To address these challenges, 
we propose the 
MPA to effectively integrate and align information from both modalities. 
MPA employs a re-embedding strategy, which first extracts the spatial Fourier components from both modalities and then enhances the channel Fourier components for each spatial Fourier component separately.
Notably, the spatial-based Fourier component serves as an efficient representation of global information, while delving into its channel-dimensional representation further enhances its descriptive power.

\begin{figure*}[!htb]
	\centering \includegraphics[width=0.9\textwidth]{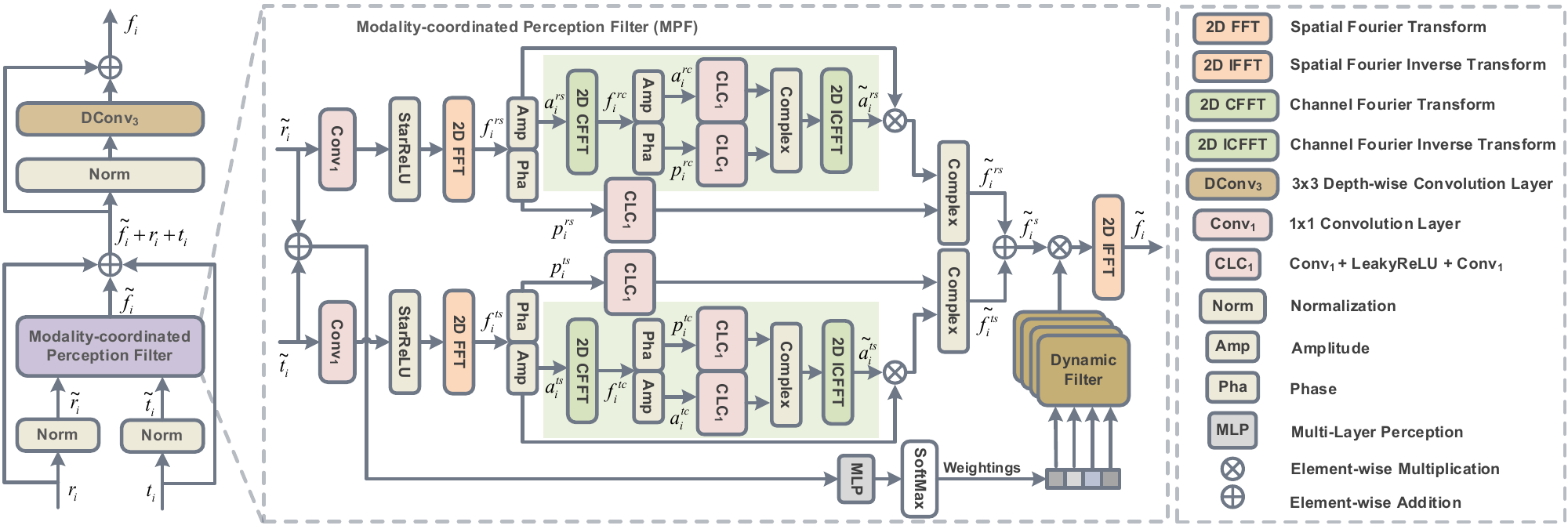}
	\caption{Architectures of the Modal-coordinated Perception Attention (MPA) and its core component Modal-coordinated Perception Filter (MPF).
    The RGB feature $r_i$ and thermal feature $t_i$ serve as inputs to the MPA, producing the output $f_i$, while $\widetilde {f_i}$ denotes the output of the MPF.}
\label{fig:mpa}
\end{figure*}

The diagram of MPA is shown in Fig. \ref{fig:mpa}, which follows a Transformer-like structure. It primarily consists of a Modal-coordinated Perception Filter (MPF), normalization layers $Norm(\cdot)$, a $3 \times 3$ depth-wise separable convolution layer $DConv_{3}(\cdot)$ in the feed-forward network, and residual connections. This process can be formulated as:
\begin{equation}
\begin{aligned}
\widetilde {{f_i}} = MPF\left( {Norm\left( {{r_i}} \right),Norm\left( {{t_i}} \right)} \right),i = 1,2,3,4,
\end{aligned}
\end{equation}
\begin{equation}
\begin{aligned}
{f_i} = DConv_3 \left( {Norm\left( {\widetilde {{f_i}} + {r_i} + {t_i}} \right)} \right) + {\widetilde {{f_i}} + {r_i} + {t_i}},
\end{aligned}
\end{equation}
where $+$ is element-wise addition. $\widetilde {{f_i}}$ and $f_i \in \mathbb{R}^{C \times H \times W}$ represent the output of MPF and MPA, respectively, with $C$, $H$, and $W$ denoting the number of channels, height, and width.

In MPF, we apply a sequence of operations to the normalized bimodal features $\widetilde {{r_i}}$ and $\widetilde {{t_i}}$: a $1 \times 1$ convolution layer $Conv_1(\cdot)$, followed by a StarReLU \cite{yu2022metaformer} layer $SR(\cdot)$, and finally, a 
FFT in the spatial dimension, to obtain the spatial Fourier components $f_i^{rs}$ and $f_i^{ts}$:
\begin{equation}
\begin{aligned}
f_i^{rs} = FFT(SR(Conv_1(\widetilde {{r_i}}))),
\end{aligned}
\end{equation}
\begin{equation}
\begin{aligned}
f_i^{ts} = FFT(SR(Conv_1(\widetilde {{t_i}})).
\end{aligned}
\end{equation}

\begin{figure}[!htp]
	\centering \includegraphics[width=0.48\textwidth]{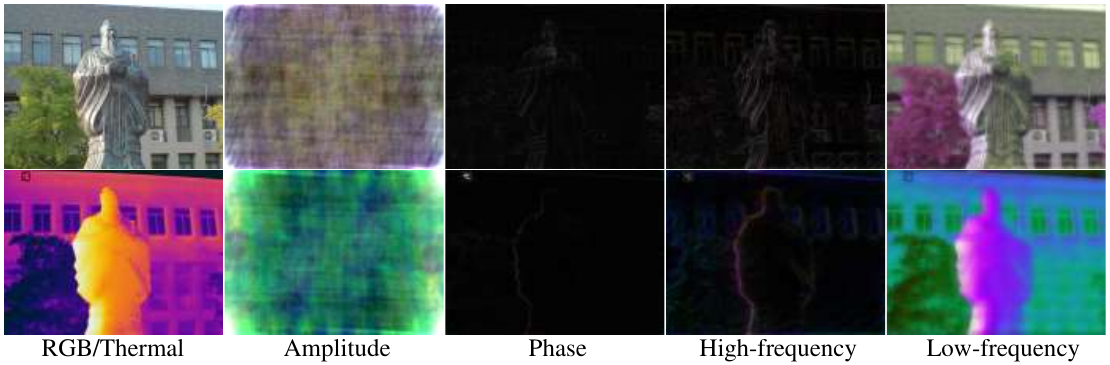}
	\caption{The visualization for the amplitude, phase, high-frequency, and low-frequency components of RGB and thermal images.}
\label{fig:visedge}
\end{figure}

To efficiently integrate bimodal features and enhance saliency representation, we decompose the spatial and channel Fourier components of each modality into amplitude and phase, and enhance them individually. As shown, Fig. \ref{fig:visedge} illustrates the frequency component decomposition of bimodal images where the amplitude captures the intensity distribution and visual saliency of each modality, aiding in the identification of overall salient regions. Enhancing the amplitude component reinforces global saliency cues and suppresses background noise. In contrast, the phase encodes fine-grained spatial details and object boundaries, which are essential for accurate edge delineation and precise localization. Enhancing the phase component refines local structural information and promotes inter-modal structural coherence. This approach enables better preservation of salient cues within each modality and effectively promotes their synergy during fusion.

First, in the channel Fourier component enhancement stage, we extract the amplitude $a_i^{rs}$ of $f_i^{rs} \in \mathbb{R}^{C \times H \times (\frac{W}{2}+1)}$ and $a_i^{ts}$ of $f_i^{ts} \in \mathbb{R}^{C \times H \times (\frac{W}{2}+1)}$ for the secondary FFT in the channel dimension (CFFT), obtaining the channel Fourier components $f_i^{rc}$ and $f_i^{tc}$:
\begin{equation}
\begin{aligned}
a_i^{rs} = Amp(f_i^{rs}),a_i^{ts} = Amp(f_i^{ts}),
\end{aligned}
\end{equation}
\begin{equation}
\begin{aligned}
f_i^{rc} = CFFT(a_i^{rs}),f_i^{tc} = CFFT(a_i^{ts}).
\end{aligned}
\end{equation}
We then enhance the amplitude and phase components using $CLC_1(\cdot)$, which consists of two $Conv_1(\cdot)$ and a LeakyReLU layer.
This process is applied to the amplitude component $a_i^{rc}$, the phase component $p_i^{rc}$ of $f_i^{rc} \in \mathbb{R}^{(\frac{C}{2}+1) \times H \times (\frac{W}{2}+1)}$ and $a_i^{tc}$, $p_i^{tc}$ of $f_i^{tc} \in \mathbb{R}^{(\frac{C}{2}+1) \times H \times (\frac{W}{2}+1)}$, respectively. 
After that, the channel-enhanced amplitude and phase components are composed into complex numbers $Complex(\cdot)$ and undergo an inverse CFFT (ICFFT). The process is denoted as:
\begin{equation}
\begin{aligned}
a_i^{rc} = Amp(f_i^{rc}),p_i^{rc} = Pha(f_i^{rc}),
\end{aligned}
\end{equation}
\begin{equation}
\begin{aligned}
\widetilde {a_i^{rs}} = ICFFT\left( {Complex\left( {CLC_1\left( {a_i^{rc}} \right),CLC_1\left( {p_i^{rc}} \right)} \right)} \right),
\end{aligned}
\end{equation}
\begin{equation}
\begin{aligned}
a_i^{tc} = Amp(f_i^{tc}),p_i^{tc} = Pha(f_i^{tc}),
\end{aligned}
\end{equation}
\begin{equation}
\begin{aligned}
\widetilde {a_i^{ts}} = ICFFT\left( {Complex\left( {CLC_1\left( {a_i^{tc}} \right),CLC_1\left( {p_i^{tc}} \right)} \right)} \right).
\end{aligned}
\end{equation}
To effectively integrate the channel-enhanced components with the original amplitude components, we utilize a shortcut connection in each branch. 
Subsequently, we combine the amplitude and phase components of the spatial Fourier representations from both branches as complex numbers. 
We then employ an element-wise addition to facilitate the complementarity of the two modalities:
\begin{equation}
\begin{aligned}
\widetilde {f_i^{rs}} = Complex\left( {\left( {a_i^{rs} \times \widetilde {a_i^{rs}}} \right),CLC_1\left( {p_i^{rs}} \right)} \right),
\end{aligned}
\end{equation}
\begin{equation}
\begin{aligned}
\widetilde {f_i^{ts}} = Complex\left( {\left( {a_i^{ts} \times \widetilde {a_i^{ts}}} \right),CLC_1\left( {p_i^{ts}} \right)} \right),
\end{aligned}
\end{equation}
\begin{equation}
\begin{aligned}
\widetilde {{f_i^{s}}} = \widetilde {f_i^{rs}} + \widetilde {f_i^{ts}},
\end{aligned}
\end{equation}
where $\times$ is element-wise multiplication.

Second, in the spatial Fourier component enhancement stage, 
we introduce a dynamic filter $DF\left(  \cdot  \right) \in \mathbb{R}^{C \times H \times (\frac{W}{2}+1)}$ \cite{tatsunami2024fft} to perform spatial filtering for the channel-enhanced fusion feature $\widetilde {{f_i^{s}}}$. $DF$ consists of $N$ learnable filters $\left\{ {{LF_1},...,{LF_N}} \right\}$ and a multi-layer perception $MLP\left(  \cdot  \right)$ is used to assign weights for learnable filters. 
Additionally, we utilize $\widetilde {{r_i}}$ and $\widetilde {{t_i}}$ to provide initial weights for filters. 
This process can be formulated as:
\begin{equation}
\begin{aligned}
\widetilde {{f_i}} = IFFT\left( {DF\left( {{\widetilde {{r_i}}},{\widetilde {{t_i}}}} \right) \times \widetilde {{f_i^s}}} \right),
\end{aligned}
\end{equation}
and
\begin{equation}
\begin{aligned}
DF{\left( {{\widetilde {{r_i}}},{\widetilde {{t_i}}}} \right)_{j,:,:}} = \sum\limits_{n = 1}^N {\left( {\frac{{{e^{{S_{\left( {j - 1} \right)N + n}}}}}}{{\sum\nolimits_{m = 1}^N {{e^{{S_{\left( {j - 1} \right)N + m}}}}} }}} \right)}{LF_n},
\end{aligned}
\end{equation}
where
\begin{equation}
\begin{aligned}
{\left( {s_1, \ldots, s_{NC}} \right)^T} = MLP\left( \frac{{\sum_{h,w} \left( {\widetilde {{r_i}} + \widetilde {{t_i}}} \right)_{:,h,w}}}{{HW}} \right),
\end{aligned}
\end{equation}
where $j$ represents the channel index, $m$ and $n$ denote filter indexes. In this paper, $N$ is set to 8 to avoid over-computing. 

\subsection{Frequency-decomposed Edge-aware Block (FEB)} 
\label{sec:FEB}
The precise identification of object details is crucial for improving detection performance. 
However, low-level features extracted from the initial layers of a network often contain boundary details alongside cluttered background noise and interference. 
This phenomenon poses a significant challenge for the decoder in making accurate predictions. 
To clarify boundary details and mitigate the impact of noise, we propose 
the FEB, a novel module specifically designed to provide the decoder with reliable edge features.

\begin{figure}[!htp]
	\centering \includegraphics[width=0.45\textwidth]{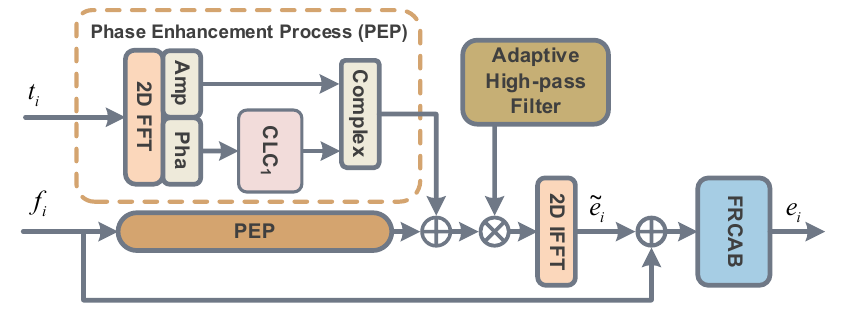}
	\caption{Architecture of the Edge Frequency Extraction Block (EFEB). 
    It enhances the low-level thermal feature $t_i$ and fusion feature $f_i$ through the Phase Enhancement Process (PEP) and an adaptive high-pass filter, thereby obtaining clear edge features $e_i$.}
\label{fig:efeb}
\end{figure}

In Fig. \ref{fig:visedge}, in addition to amplitude reflecting intensity distribution and phase encoding spatial details as discussed in Sec. \ref{sec:MPA}, we also observe that low-frequency components primarily convey style and content information, while high-frequency components encapsulate detailed edges and textures.
Furthermore, 
RGB images' phase and high-frequency components contain more complex background textures together with boundary details, 
whereas those of thermal images exhibit clearer object contours and simpler textures \cite{10127616,chen2022modality,9611276,sun2023catnet}.
Building on this observation, we develop a key component of the FEB, namely the Edge Frequency Extraction Block (EFEB), as illustrated in Fig. \ref{fig:efeb}. 

The FEB contains two EFEBs. 
Considering that RGB features contain more background texture noise, EFEBs 
introduce
low-level thermal and fusion features $\left\{ {{t_i},{f_i}} \right\}_{i = 1}^2$ undergoing a Phase Enhancement Process (PEP) to highlight object boundaries, followed by integration. 
An adaptive high-pass filter $HF$ is then employed to retain detailed edge information while removing style information. Finally, the output is refined by a Fourier Residual Channel Attention Block (FRCAB) to mitigate the network's preference for low-frequency information, with details provided in Sec. \ref{sec:FRCAB}. 

In PEP, a given input $x$ undergoes a 
FFT in the spatial dimension initially, followed by phase component enhancement. The PEP $PEP(\cdot)$ can be mathematically expressed as:
\begin{equation}
\begin{aligned}
a^{xs} = Amp(FFT(x)),p^{xs} = Pha(FFT(x)),
\end{aligned}
\end{equation}
\begin{equation}
\begin{aligned}
\widetilde {{x}} = Complex({a^{xs}},CLC_1\left( {{p^{xs}}} \right)).
\end{aligned}
\end{equation}
Then, EFEB can then be formulated as:
\begin{equation}
\begin{aligned}
\widetilde {{e_i}} = IFFT((PEP(t_i) + PEP(f_i)) \times HF),
\end{aligned}
\end{equation}
\begin{equation}
\begin{aligned}
{e_i} = FRCAB ( \widetilde {{e_i}} + f_i ),i = 1,2.
\end{aligned}
\end{equation}
Finally, the output edge feature $e_3$ of FEB is calculated as:
\begin{equation}
\begin{aligned}
{e_3} = DNRU\left( {Concat\left( {{e_1},UP_2\left( {{e_2}} \right)} \right)} \right),
\label{eq:21}
\end{aligned}
\end{equation}
where $UP_2(\cdot)$ represents 2$\times$ up-sampling, and $Concat\left(  \cdot  \right)$ denotes concatenation. $DNRU(\cdot)$ represents a set of operations including $DConv_{3}(\cdot)$, $Norm(\cdot)$, a ReLU layer, and $UP_2(\cdot)$.

\subsection{Fourier Residual Channel Attention Block (FRCAB)} \label{sec:FRCAB}
Due to the inherent bias of neural networks, low-resolution encoder features tend to emphasize low-frequency semantic information. However, for dense prediction tasks, the decoding stage should focus more on high-frequency components to generate precise high-resolution saliency maps. Additionally, as the channel dimensions are progressively reduced, treating all feature channels equally can hinder the generation of accurate saliency maps. To address these issues, we draw inspiration from the Residual Channel Attention Block (RCAB) \cite{zhang2018image} and propose the 
FRCAB, which is strategically placed at each layer of the decoding stage. 

\begin{figure}[!htp]
	\centering \includegraphics[width=0.47\textwidth]{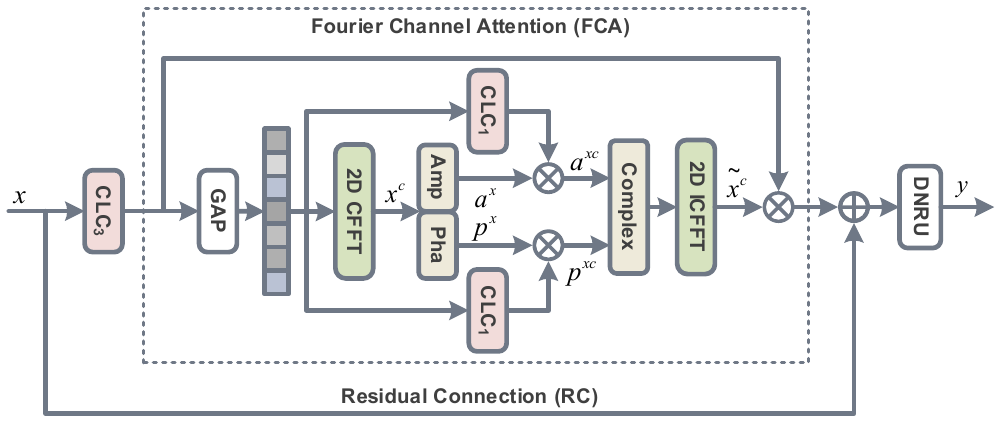}
	\caption{Architecture of the Fourier Residual Channel Attention Block (FRCAB).
    It mainly consists of a Residual Connection (RC) and a Fourier Channel Attention (FCA), where $x$ and $y$ denote its input and output, respectively.}
\label{fig:frcab}
\end{figure}

The FRCAB integrates two key components: a Residual Connection (RC) \cite{zhang2018image} and Fourier Channel Attention (FCA), as shown in Fig. \ref{fig:frcab}. The RC enables the decoder to focus more on high-frequency information, while FCA extracts inter-channel statistics in the Fourier domain to further enhance the model's capability to discriminate global information. 

As illustrated in Fig. \ref{fig:frcab}, given an input feature $x \in \mathbb{R}^{C \times H \times W}$, 
the FRCAB operates as follows:
First, a global vector is extracted using global average pooling $GAP(\cdot)$ after processing $x$ with a $CLC_3(\cdot)$ with $3 \times 3$ kernel size.
Then, a CFFT is performed and $CLC_1(  \cdot)$ operations are used to enhance the amplitude $a^x \in \mathbb{R}^{(\frac{C}{2}+1) \times 1 \times 1}$ and phase $p^x \in \mathbb{R}^{(\frac{C}{2}+1) \times 1 \times 1}$ components of $x$. 
Subsequently, an ICFFT is performed, and two residual connections are constructed. 
Finally, we employ a $DNRU(\cdot)$ to reduce the number of channels and up-sampling features, and obtain the output $y$. The FRCAB $FRCAB(\cdot)$ can be mathematically described as:
\begin{equation}
\begin{aligned}
{x^c} = CFFT(GAP\left( {CLC_3\left( x \right)} \right)),
\end{aligned}
\end{equation}
\begin{equation}
\begin{aligned}
a^{x} = Amp(x^{c}),p^{x} = Pha(x^{c}),
\end{aligned}
\end{equation}
\begin{equation}
\begin{aligned}
{a^{xc}} = {a^x} \times CLC_1 ( GAP\left( {CLC_3\left( x \right)} \right) ),
\end{aligned}
\end{equation}
\begin{equation}
\begin{aligned}
{p^{xc}} = {p^x} \times CLC_1 ( GAP\left( {CLC_3\left( x \right)} \right) ),
\end{aligned}
\end{equation}
\begin{equation}
\begin{aligned}
\widetilde {{x^c}} = ICFFT\left( {Complex\left( {{a^{xc}},{p^{xc}}} \right)} \right),
\end{aligned}
\end{equation}
\begin{equation}
\begin{aligned}
y = DNRU( \widetilde {{x^c}} \times CLC_3(x) + x).
\end{aligned}
\end{equation}

After applying the FRCAB in the decoding stage, guided by the clear edge features $e_3$ 
generated by the 
FEB (Sec. \ref{sec:FEB}), 
the decoding process proceeds as follows to generate the final saliency map $\mathcal{S}$:
\begin{equation}
\begin{aligned}
d_4 = FRCAB(f_4),
\end{aligned}
\end{equation}
\begin{equation}
\begin{aligned}
d_3 = FRCAB(Concat(f_3, d_4)),
\end{aligned}
\end{equation}
\begin{equation}
\begin{aligned}
d_2 = FRCAB(Concat(f_2, d_3,UP_2(d_4))),
\end{aligned}
\end{equation}
\begin{equation}
\begin{aligned}
d_1 = FRCAB(Concat(f_1, d_2, UP_2(d_3), UP_4(d_4))),
\end{aligned}
\end{equation}
\begin{equation}
\begin{aligned}
\mathcal{S} = DNRU(Concat(d_1, e_3)),
\end{aligned}
\end{equation}
where $\left\{ {{d_i}} \right\}_{i = 1}^4$ and $\left\{ {{f_i}} \right\}_{i = 1}^4$ stand for decoder features and fusion features obtained by the 
MPA (Sec. \ref{sec:MPA}).

\subsection{Bi-domain Learning}
\label{sec:bidomain}
We trained our \proposed with various learning strategies in the spatial and frequency domains, aiming to generate accurate saliency maps and reliable edge features.

\emph{1) Frequency Domain Learning:} As the model deepens, important visual details can be lost, and spectral bias may hinder the model's ability to learn high-accuracy edge features.
To address these problems, 
we propose Co-focus Frequency Loss (CFL), designed to provide effective guidance for FEB (Sec. \ref{sec:FEB}). 
CFL leverages 
FFT to measure distances in the Fourier domain while cross-referencing details in the original feature space.
This allows for the optimization of global information and aids FEB in addressing difficult frequencies. 

For every pixel $(u,v)$ in the frequency domain, CFL is defined as:
\begin{equation}
\begin{aligned}
{{\cal L}_{CFL}} = \frac{1}{{HW}}\sum\limits_{u = 0}^{H - 1} {\sum\limits_{v = 0}^{W - 1} {{w}{{\left| {FFT\left( {\mathcal{E}} \right) - FFT\left( E \right)} \right|}^2}}}, 
\end{aligned}
\end{equation}
where
\begin{equation}
\begin{aligned}
\mathcal{E} = DConv_3({e_3}),
\end{aligned}
\end{equation}
and $E$ denotes the edge of objects in ground-truth $G$, obtained by the Canny operator \cite{canny1986computational}. $e_3$ is the output of FEB and $\mathcal{E}$ is the predicted edge map.

To focus learning on hard frequencies, we introduce the Co-focus Frequency Matrix (CFM) $w$ to enhance their associated weights.
As discussed in Sec. \ref{sec:FEB}, the phase component contains plentiful edge frequencies. Thus, 
while adopting $r_1$ and $t_1$ as co-directors, 
we set the amplitude components of $r_1$, $t_1$ to the mean values $Mean(\cdot)$ to emphasize their phase components:
\begin{equation}
\begin{aligned}
\mathcal{R} = UP_4({DConv_3({r_1})}), \mathcal{T} = UP_4({DConv_3({t_1})}),
\end{aligned}
\end{equation}
\begin{equation}
\begin{aligned}
{\widetilde {{\mathcal{R}}}} = Complex(Mean\left( {Amp\left( {FFT\left( {{\mathcal{R}}} \right)} \right)} \right),Pha\left( {FFT\left( {{\mathcal{R}}} \right)} \right)),
\end{aligned}
\end{equation}
\begin{equation}
\begin{aligned}
{\widetilde {{\mathcal{T}}}} = Complex(Mean\left( {Amp\left( {FFT\left( {{\mathcal{T}}} \right)} \right)} \right),Pha\left( {FFT\left( {{\mathcal{T}}} \right)} \right)).
\end{aligned}
\end{equation}
Then, CFM is expressed as:
\begin{equation}
\begin{aligned}
\label{Eq:alpha}
    w = & (| {\widetilde {{\mathcal{R}}}} - FFT(E) | + | {\widetilde {{\mathcal{T}}}} - FFT(E) | \\
    & + | FFT(\mathcal{E}) - FFT(E) |)^\alpha,
\end{aligned}
\end{equation}
where $\alpha$ is a scaling factor. The introduction of $r_1$ and $t_1$ not only provides raw frequency information but also enhances the representation of edge information in the original feature space of both modalities.

\emph{2) Spatial Domain Learning:} 
We also employ ${{\cal L}_{BCE}}$ to supervise the learning of the edge map $\mathcal{E}$ in the spatial domain:
\begin{equation}
\begin{aligned}
{{\cal L}_E} = {{\cal L}_{BCE}}\left( {\mathcal{E}, E} \right).
\end{aligned}
\end{equation}
This bi-domain learning strategy ensures obtaining the high-quality edge map.

In addition, to alleviate network bias and ensure the accuracy of the features in the decoding stage, we construct a multi-level loss with ${{\cal L}_{BCE}}$ and ${{\cal L}_{IoU}}$:
\begin{equation}
\begin{aligned}
    {\cal L}_{D} = \sum\limits_{i = 1}^4 \biggl( & {{\cal L}_{BCE}\bigl( UP\bigl(DConv_3(d_i) \bigr), G \bigr)} \\
    & + {{\cal L}_{IoU}\bigl( UP\bigl( DConv_3(d_i) \bigr), G \bigr)},
\end{aligned}
\end{equation}
where $\left\{ {{d_i}} \right\}_{i = 1}^4$ are decoder features generated by FRCABs.

Similarly, the saliency loss is defined as:
\begin{equation}
\begin{aligned}
{{\cal L}_S} = {{\cal L}_{BCE}}\left( {\mathcal{S}, G} \right) + {{\cal L}_{IoU}}\left( {\mathcal{S}, G} \right).
\end{aligned}
\end{equation}

In summary, the total loss ${\cal L}_{total}$ for training is denoted as:
\begin{equation}
\begin{aligned}
{{\cal L}_{total}} = {\lambda _1}{{\cal L}_S} + {\lambda _2}{{\cal L}_{D}} + {\lambda _3}{{\cal L}_{E}} + {\lambda _4}{{\cal L}_{CFL}},
\label{eq:total_loss}
\end{aligned}
\end{equation}
where $\left\{ {{\lambda_i}} \right\}_{i = 1}^4$ denote weight parameters corresponding to different losses. In this paper, we set them to 1.

\section{Experiments}
\subsection{Datasets}
To evaluate the effectiveness of our proposed \proposed, we first conducted experiments on the RGB-T SOD task. To further evaluate its generalization, we extended the comparison to RGB-D-T and RGB-D SOD tasks.

\emph{1) RGB-T SOD:} We conducted experiments on the four RGB-T SOD benchmark datasets, namely VT821 \cite{wang2018rgb}, VT1000 \cite{8744296}, VT5000 \cite{9767629}, and VI-RGBT1500 \cite{10003255}. The VT821 dataset consists of 821 pairs of RGB-T images captured under different environmental conditions.
The VT1000 dataset comprises 1000 aligned RGB-T images with detailed annotations tailored for evaluation, providing a reliable benchmark for performance comparison. The VT5000 dataset offers 5000 pairs of high-resolution, low-bias RGB-T images that cover a broad range of challenging scenes. The VI-RGBT1500 dataset consists of 1500 RGB-T image pairs captured under different illuminations to validate the benefits of incorporating thermal modality. 
Following \cite{10003255,10127616}, we used 2500 RGB-T image pairs from the VT5000 dataset as our first training set, while the remaining images and those from the VT821, VT1000, and VI-RGBT1500 datasets served as the testing set.

\emph{2) RGB-D-T SOD:} 
We employed the VDT-2048 dataset \cite{HWSI} for the experiments. It provides 2048 aligned RGB-depth-thermal image pairs for robot visual awareness under various light conditions. Following \cite{HWSI}, we employed 1048 image pairs from the VDT-2048 dataset as the training set, and the remaining 1000 image pairs formed the testing set.

\emph{3) RGB-D SOD:} 
We performed comparison experiments on five RGB-D SOD benchmark datasets, namely NLPR \cite{NLPR}, NJUD \cite{NJUD}, DUT-RGBD \cite{DUT-RGBD}, SIP \cite{SIP}, and STERE \cite{STERE}.
The NLPR dataset consists of 1000 RGB-D image pairs containing diverse indoor-outdoor natural scenes.
The NJUD dataset contains 1985 image pairs generated from Internet videos and images.
The DUT-RGBD dataset contains 1200 RGB-D image pairs from complex real-world scenes.
The SIP dataset collects 929 high-definition RGB-D image pairs of human life scenes.
The STERE dataset consists of 1000 Internet RGB-D image pairs and corresponding manually labeled masks.
Following \cite{PICRNet,liu2024vst++,10015667}, we conducted two training setups: (1) using 700 image pairs from the NLPR dataset and 1485 from the NJUD dataset, and (2) extending the first setup with an additional 800 image pairs from the DUT-RGBD dataset.

\subsection{Evaluation Metrics}
To comprehensively evaluate the results of each experiment, we employed six widely used evaluation metrics. The E-measure \cite{ijcai2018p97} ($Em$) considers both image-level statistics and local pixel differences between the predictions and the ground-truths.
The S-measure \cite{8237749} ($Sm$) evaluates the structural similarity between the predictions and the ground-truths. The F-measure \cite{5206596} (${F_\beta}$) is a comprehensive metric, calculated by weighted harmonic averaging on precision and recall. 
We also used the weighted F-measure \cite{6909433} (${wF_\beta}$) to complement the F-measure, addressing issues such as interpolation defects and dependency defects. 
\begin{table}[!htp]
  \centering
  \fontsize{8}{10}\selectfont
  \renewcommand{\arraystretch}{1}
  \renewcommand{\tabcolsep}{0.8mm}
  \scriptsize
  \caption{Selection of Scaling Factor $\alpha$ in Co-focus Frequency Loss. The Best Results are Labeled \textcolor{red}{\textbf{Red}}.}
\label{tab:cfl_hyperparam}
  \scalebox{1}{
  \begin{tabular}{c|cc|cc|cc|cc}
  \hline\toprule
    \multirow{2}{*}{\centering Settings}  & \multicolumn{2}{c|}{\centering VT821} & \multicolumn{2}{c|}{\centering VT1000} & \multicolumn{2}{c|}{\centering VT5000} &\multicolumn{2}{c}{\centering VI-RGBT1500}\\
   \multicolumn{1}{c|}{} & $F_\beta\uparrow$ & $\mathcal{M}\downarrow$ & $F_\beta\uparrow$ & $\mathcal{M}\downarrow$ & $F_\beta\uparrow$ & $\mathcal{M}\downarrow$ & $F_\beta\uparrow$ & $\mathcal{M}\downarrow$ \\
\midrule
    $\alpha=0.5$ & 0.868 &	0.024 & 0.899 & 0.015 & 0.873 & 0.025 & 0.885 & 0.028\\			
    $\alpha=1.0$ & \textcolor{red}{\textbf{0.872}} & \textcolor{red}{\textbf{0.023}} & \textcolor{red}{\textbf{0.903}} & \textcolor{red}{\textbf{0.014}} & \textcolor{red}{\textbf{0.884}} & \textcolor{red}{\textbf{0.023}} & 0.893 & \textcolor{red}{\textbf{0.026}}\\			
    $\alpha=1.5$ & 0.865 & 0.024 & 0.901 & \textcolor{red}{\textbf{0.014}} & 0.878 & 0.024 & \textcolor{red}{\textbf{0.897}} & \textcolor{red}{\textbf{0.026}} \\		
    $\alpha=2.0$ & 0.860 & 0.026 &	0.894 & 0.016 & 0.871 & 0.025 & 0.880 & 0.029\\
    \bottomrule
    \hline
  \end{tabular}}
\end{table}
Additionally, we calculated the Mean absolute error \cite{6247743} ($\mathcal{M}$), which represents the average discrepancy between each pixel in the prediction and the ground-truth. Finally, Precision-recall (PR) curve \cite{10015667} is drawn based on precision-recall pairs, where curves closer to $\left( {1,1} \right)$ indicate better model performance.

\subsection{Implementation Details}
The backbone of \proposed was initialized with the training parameters of CDFFormer-M \cite{tatsunami2024fft}, while the remaining parameters were initialized using 
Kaiming initialization \cite{he2015delving}.
The model was trained on an NVIDIA Tesla P100 GPU with 16GB of memory using Adam optimization \cite{kingma2014adam} and a batch size of 3. Training lasted for 240 epochs, starting with an initial learning rate of $2 \times 10^{-5}$, which decayed to one-tenth every 80 epochs. We set the input size to $384 \times 384$, and performed a series of data augmentation operations like random cropping, horizontal flipping, and multi-angle rotation to prevent overfitting.

The hyperparameters of CFL, specifically the scaling factor $\alpha$ in Eq. (\ref{Eq:alpha}), were determined through careful testing. We fixed $\alpha=1$ for our experiments, unless stated otherwise. The representative values of $\alpha$ 
are presented in Table \ref{tab:cfl_hyperparam}.

\begin{table*}[!htp]
  \centering
  \fontsize{8}{10}\selectfont
  \renewcommand{\arraystretch}{1.1}
  \renewcommand{\tabcolsep}{0.35mm}
  \scriptsize
  \caption{Comparison of E-measure ($Em$), S-measure ($Sm$), F-measure (${F_\beta}$), Weighted F-measure (${wF_\beta}$), and Mean Absolute Error ($\mathcal{M}$) Between Our \proposed and State-of-the-Art RGB-T SOD Models on the VT821, VT1000, VT5000, and VI-RGBT1500 datasets. The Best Results are Labeled \textcolor{red}{\textbf{Red}} and the Second Best Results are Labeled \textcolor{blue}{Blue}. FPS Was Tested on an NVIDIA RTX4060 Laptop GPU.}
\label{tab:RGBTcomparison}
  \scalebox{0.85}{
  \begin{tabular}{c|c|ccccc|ccccc|ccccc|ccccc|c|c|c}
  \hline\toprule
   \multirow{2}{*}{\centering Model} & \multirow{2}{*}{\centering Backbone} & \multicolumn{5}{c|}{\centering VT821} & \multicolumn{5}{c|}{\centering VT1000} & \multicolumn{5}{c|}{\centering VT5000} & \multicolumn{5}{c|}{\centering VI-RGBT1500} & \multirow{2}{*}{\centering Param(M)} & \multirow{2}{*}{\centering FLOPs(G)} &\multirow{2}{*}{\centering FPS}\\
   & & $Em\uparrow$ & $Sm\uparrow$ & $wF_{\beta} \uparrow$ & $F_{\beta}\uparrow$ & $\mathcal{M}\downarrow$ &$Em\uparrow$ & $Sm\uparrow$ &$wF_{\beta} \uparrow$ & $F_{\beta}\uparrow$ & $\mathcal{M}\downarrow$ & $Em\uparrow$ & $Sm\uparrow$ & $wF_{\beta} \uparrow$ & $F_{\beta}\uparrow$ & $\mathcal{M}\downarrow$ & $Em\uparrow$ & $Sm\uparrow$ & $wF_{\beta} \uparrow$ & $F_{\beta}\uparrow$ & $\mathcal{M}\downarrow$ & & &\\
\midrule
    \multicolumn{25}{c}{\centering CNN-based Model}\\
\midrule
    CSRNet$_{21}$ \cite{9505635} & ESPNetv2 & 0.908 & 0.885 & 0.821 & 0.830 & 0.038 & 0.925 & 0.918 & 0.878 & 0.877 & 0.024 & 0.905 & 0.868 & 0.797 & 0.811 & 0.042 & 0.836 & 0.715 & 0.572 & 0.661 & 0.089 & 4.6 & 4.2 & 40.2\\	
    MIDD$_{21}$ \cite{9454273} & VGG-16 & 0.895 &0.871 & 0.760 & 0.804 & 0.045 & 0.933 & 0.915 & 0.856 & 0.822 & 0.027 & 0.897 & 0.868 & 0.763 & 0.801 & 0.043 & 0.870 & 0.762 & 0.665 & 0.765 & 0.069 & 52.4 & 257.9 & 86.3\\	
    OSRNet$_{22}$ \cite{9803225} & VGG-16 & 0.896 & 0.875 & 0.801 & 0.813 & 0.043 & 0.935 & 0.926 & 0.891 & 0.892 & 0.022 & 0.908 & 0.875 & 0.807 & 0.823 & 0.040 & 0.917 & 0.872 & 0.811 & 0.836 & 0.044 &42.4 & 15.6 & 95.6\\		
    ADF$_{23}$ \cite{9767629} & VGG-16 & 0.842 & 0.810 & 0.626 & 0.716 & 0.077 & 0.921 & 0.910 & 0.804 & 0.847 & 0.034 & 0.891 & 0.864 & 0.722 & 0.778 & 0.048 & 0.888 & 0.831 & 0.664 & 0.777 & 0.074 &83.1 & 77.6 & 7.3\\
    TNet$_{23}$ \cite{9926193} & ResNet-50 & 0.919 & 0.899 & 0.841 & 0.842 & 0.030 & 0.937 & 0.929 & 0.895 & 0.889 & 0.021 & 0.927 & 0.895 & 0.840 & 0.846 & 0.033 & 0.943 & 0.894 & 0.855 & 0.869 & 0.034 & 87.0 & 39.7 & 59.0\\
    LSNet$_{23}$ \cite{10042233} & MobileNetv2 & 0.911 & 0.878 & 0.809 & 0.825 & 0.033 & 0.935 & 0.925 & 0.887 & 0.885 & 0.023 & 0.915 & 0.877 & 0.806 & 0.825 & 0.037 & 0.935 & 0.883 & 0.835 & 0.859 & 0.037 & 5.4 & 1.2 & 88.4\\
    MGAI$_{23}$ \cite{10003255} & ResNet-50 & 0.913 & 0.891 & 0.824 & 0.829 & 0.031 & 0.935 & 0.929 & 0.893 & 0.885 & 0.021 & 0.915 & 0.883 & 0.815 & 0.824 & 0.034 & 0.938 & 0.881 & 0.834 & 0.862 & 0.038 & 86.8 & 94.2 & 4.7\\
    LAFB$_{24}$ \cite{wang2024learning} & Res2Net-50 & 0.915 & 0.884 & 0.817 & 0.842 & 0.034 & 0.945 & 0.932 & 0.905 & 0.905 & 0.018 & 0.931 & 0.893 & 0.841 & 0.857 & 0.030 & 0.937 & 0.887  & 0.847 & 0.867  & 0.035 & 453.0 & 139.7 & 29.3\\
\midrule
    \multicolumn{25}{c}{\centering Transformer-based Model}\\
\midrule
    SwinNet$_{22}$ \cite{9611276} & Swin-B & 0.926 & 0.904 & 0.818 & 0.847 & 0.030 & 0.947 & 0.938 & 0.894 & 0.896 & 0.018 & 0.942 & 0.912 & 0.846 & 0.865 & 0.026 & 0.948 & 0.901 & 0.863 & 0.882 & 0.031 & 198.7 & 124.3 & 25.8\\
    IFFNet$_{23}$ \cite{10015881} & Segformer & 0.918 & 0.907 & 0.849 & 0.848 & 0.029 & 0.947 & 0.938 & 0.912 & 0.900 & 0.017 & 0.938 & 0.905 & 0.856 & 0.864 & 0.028 & - & - & - & - & - & - & - & -\\
    HRTNet$_{23}$ \cite{9869666} & HRFormer & 0.929 & 0.906 & 0.849 & 0.853 & 0.026 & 0.945 & 0.938 & 0.913 & 0.900 & 0.017 & 0.945 & 0.912 & 0.870 & 0.871 & 0.025 & 0.947 & 0.899 & 0.870 & 0.892 & 0.030 & 58.9 & 17.3 & 11.5\\
    CAVER$_{23}$ \cite{10015667} & ResNet-101 & 0.929 & 0.898 & 0.846 & 0.854 & 0.026 & 0.949 & 0.938 & 0.912 & 0.906 & 0.016 & 0.935 & 0.900 & 0.849 & 0.856 & 0.028 & 0.949 & 0.900 & 0.868 & 0.881 & 0.029 & 93.8 & 63.9 & 16.2\\
    CAFCNet$_{24}$ \cite{jin2024cafcnet} & ResNet-50 & 0.927 & 0.891 & 0.845 & 0.856 & 0.028 & 0.951 & 0.935 & 0.915 & \textcolor{blue}{0.913} & 0.017 & 0.939 & 0.900 & 0.858 & 0.869 & 0.027 & 0.947 & 0.894  & 0.866 & 0.881  & 0.030 & 185.7 & 324.2 & 13.8 \\
    VST++$_{24}$ \cite{liu2024vst++} & Swin-B & 0.921 & 0.904 & 0.850 & 0.848 & 0.030 & 0.951 & 0.943 & 0.920 & 0.908 & 0.016 & 0.945 & 0.915 & 0.873 & 0.875 & 0.025 & - & -  & - & -  & - & 217.7 & - & -\\
    MAGNet$_{24}$ \cite{MAGNet} & SMT & 0.930 & 0.909 & 0.863 & 0.865 & 0.026 & 0.949 & 0.938 & 0.917 & 0.912 & 0.016 & 0.943 & 0.909 & 0.870 & 0.876 & 0.025 & 0.949 & 0.899 & 0.873 & 0.885 & 0.029 & 16.1 & 9.9 & 28.1\\
    PATNet$_{24}$ \cite{PATNet} & P2T-B & 0.935 & \textcolor{blue}{0.914} & \textcolor{blue}{0.872} & 0.870 & \textcolor{blue}{0.024} & 0.952 & 0.942 & 0.920 & 0.910 & \textcolor{blue}{0.015} & 0.947 & \textcolor{blue}{0.918} & \textcolor{red}{\textbf{0.879}} & 0.883 & \textcolor{red}{\textbf{0.023}} & 0.948 & 0.908 & \textcolor{blue}{0.878} & 0.885 & 0.029 & 94.8 & 51.1 & 11.8\\
    UniTR$_{24}$ \cite{UniTR} & Swin-B & \textcolor{blue}{0.936} & 0.902 & 0.863 & \textcolor{red}{\textbf{0.875}} & 0.025 & \textcolor{blue}{0.958} & 0.939 & \textcolor{blue}{0.925} & \textcolor{red}{\textbf{0.926}} & \textcolor{red}{\textbf{0.014}} & \textcolor{blue}{0.950} & 0.909 & \textcolor{blue}{0.878} & \textcolor{red}{\textbf{0.892}} & \textcolor{red}{\textbf{0.023}} & \textcolor{blue}{0.951} & 0.895 & 0.874 & \textcolor{red}{\textbf{0.898}} & 0.029 & 146.3 & 135.0 & 31.0\\  
\midrule
    \multicolumn{25}{c}{\centering Frequency-based Model}\\
\midrule
    WaveNet$_{23}$ \cite{10127616} & WaveMLP & 0.929 & 0.912 & 0.863 & 0.857 & \textcolor{blue}{0.024} & 0.952 & \textcolor{blue}{0.945} & 0.921 & 0.909 & \textcolor{blue}{0.015} & 0.940 & 0.912 & 0.865 & 0.867 & 0.026 & \textcolor{blue}{0.951} & \textcolor{blue}{0.909} & 0.874 & 0.879 & \textcolor{blue}{0.027} & 80.7 & 64.0 & 5.3\\
\midrule
    \proposed (Ours) & CDFFormer-M & \textcolor{red}{\textbf{0.939}} & \textcolor{red}{\textbf{0.917}} & \textcolor{red}{\textbf{0.873}} & \textcolor{blue}{0.872} & \textcolor{red}{\textbf{0.023}} & \textcolor{red}{\textbf{0.959}} & \textcolor{red}{\textbf{0.948}} & \textcolor{red}{\textbf{0.928}} & 0.903 & \textcolor{red}{\textbf{0.014}} & \textcolor{red}{\textbf{0.951}} & \textcolor{red}{\textbf{0.920}} & \textcolor{red}{\textbf{0.879}} & \textcolor{blue}{0.884} & \textcolor{red}{\textbf{0.023}} & \textcolor{red}{\textbf{0.956}} & \textcolor{red}{\textbf{0.916}} & \textcolor{red}{\textbf{0.885}} & \textcolor{blue}{0.893} & \textcolor{red}{\textbf{0.026}} & 149.6 & 139.5 & 17.1\\
    \bottomrule
    \hline
  \end{tabular}}
\end{table*}

\begin{figure*}[!htp]
	\centering \includegraphics[width=0.9\textwidth]{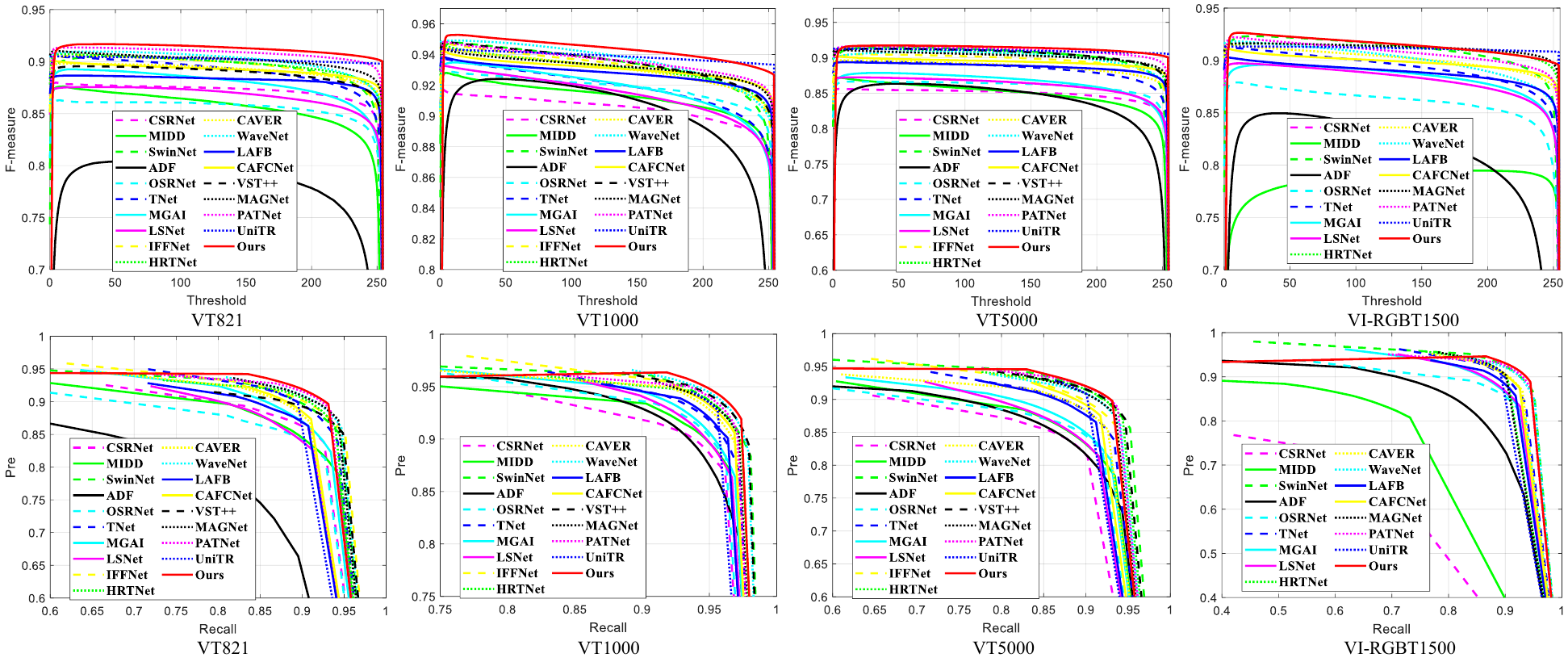}
	\caption{F-measure curves and precision-recall (PR) curves of the proposed \proposed and eighteen existing RGB-T SOD models on the VT821, VT1000, VT5000, and VI-RGBT1500 datasets.}
\label{fig:FTPR}
\end{figure*}

\subsection{Model Comparison}
To validate the effectiveness of the proposed \proposed, we conducted a comprehensive comparison with eighteen state-of-the-art (SOTA) RGB-T SOD models, namely CSRNet \cite{9505635}, MIDD \cite{9454273}, OSRNet \cite{9803225}, SwinNet \cite{9611276}, ADF \cite{9767629}, TNet \cite{9926193}, MGAI \cite{10003255}, LSNet \cite{10042233}, IFFNet \cite{10015881}, HRTNet \cite{9869666}, CAVER \cite{10015667}, WaveNet \cite{10127616}, LAFB \cite{wang2024learning}, CAFCNet \cite{jin2024cafcnet}, VST++ \cite{liu2024vst++}, MAGNet \cite{MAGNet}, PATNet \cite{PATNet}, and UniTR \cite{UniTR}. To ensure a fair comparison, all saliency maps used for the model comparison were acquired from the corresponding papers' homepages or by running open-source code. 

\emph{1) Quantitative Evaluation:} Table \ref{tab:RGBTcomparison} and Fig. \ref{fig:FTPR} present the quantitative comparison of our \proposed with eighteen existing models using six evaluation metrics: $Em$, $Sm$, $wF_\beta$, $F_\beta$, $\mathcal{M}$, and PR curve. 
In summary, \proposed outperformed eighteen existing models across all six metrics on all four benchmark datasets.
Specifically, our \proposed surpassed the previously top-performing UniTR \cite{UniTR} on the VT821, VT1000, VT5000 and VI-RGBT1500 datasets, achieving average improvements of 0.3$\%$, 2.4$\%$, 0.7$\%$, 5.5$\%$ in $Em$, $Sm$, $wF_\beta$, and $\mathcal{M}$, respectively.
Furthermore, when compared to the previous best frequency-based model, WaveNet \cite{10127616}, the average improvements in $Em$, $Sm$, $wF_\beta$, $F_\beta$, and $\mathcal{M}$ is 0.9$\%$, 0.6$\%$, 1.2$\%$, 1.1$\%$, and 6.5$\%$, respectively.
This advantage can be attributed to the design of our \proposed, which enables superior global information discrimination. Meanwhile, the proposed 
CFL provides support in optimizing global frequencies and resolving hard frequency problems.

In addition, the F-measure and PR curves in Fig. \ref{fig:FTPR} further validate the leading performance of our \proposed. In the $1^{st}_{}$ row of Fig. \ref{fig:FTPR}, F-measure curves of \proposed demonstrate the highest ${F_\beta}$ across various thresholds. Moreover, PR curves of \proposed are closer to the point with coordinates (1,1) compared to those of other models. These findings indicate \proposed' superior balance between precision and recall.

\begin{figure*}[!htp]
	\centering \includegraphics[width=1\textwidth]{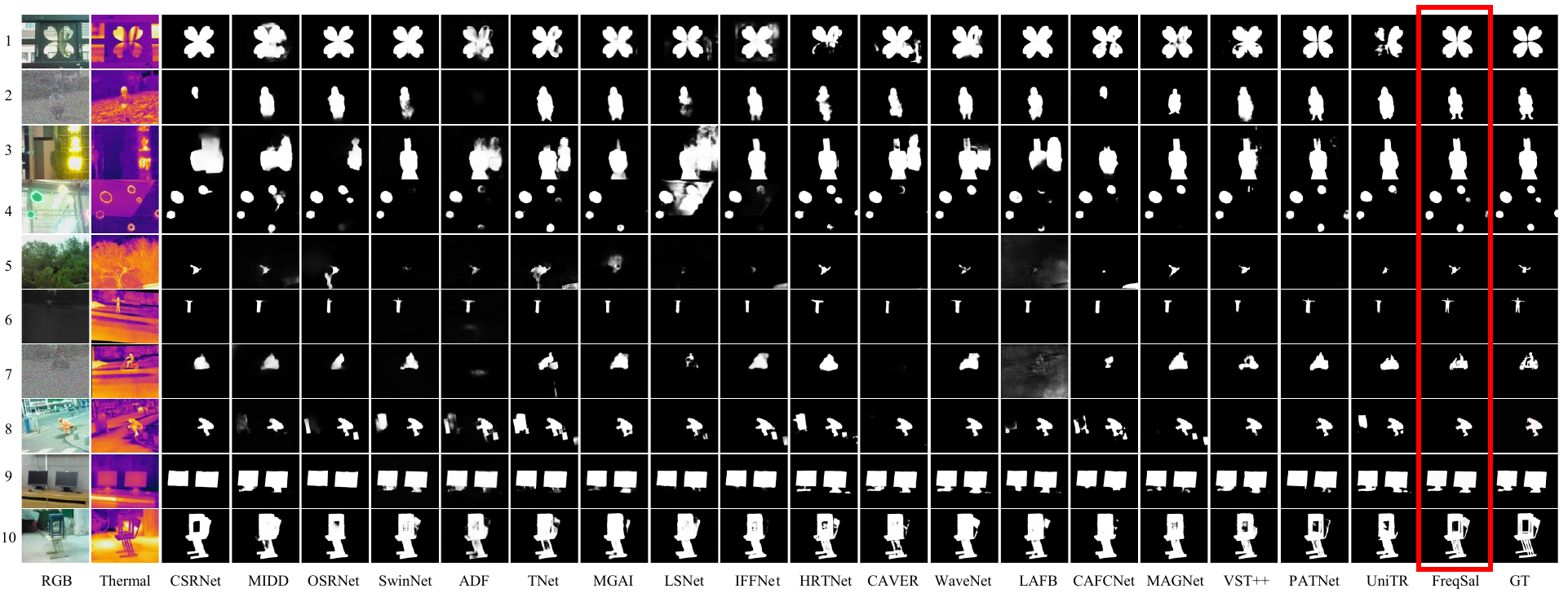}
	\caption{Visual comparison of our \proposed with eighteen existing RGB-T SOD models in various complex scenarios, including large objects ($1^{st}_{}$-$3^{rd}_{}$, $10^{th}_{}$ rows), small objects ($5^{th}_{}$-$8^{th}_{}$ rows), multiple objects ($4^{th}_{}$ and $9^{th}_{}$ rows), complex background ($1^{st}_{}$-$5^{th}_{}$, $8^{th}_{}$ and $10^{th}_{}$ rows), low light or contrast ($3^{rd}_{}$-$6^{th}_{}$ rows), small thermal difference ($2^{nd}_{}$-$3^{rd}_{}$, $5^{th}_{}$, $7^{th}_{}$, and $9^{th}_{}$ rows), low-quality RGB images ($2^{nd}_{}$ and $7^{th}_{}$ rows), and fine-grained objects ($7^{th}_{}$, and $10^{th}_{}$ rows). GT stands for ground-truth.}
\label{fig:RGBTVisualCompare}
\end{figure*}

\emph{2) Qualitative Evaluation:} Fig. \ref{fig:RGBTVisualCompare} illustrates the saliency maps generated by our proposed \proposed and representative RGB-T SOD models. It is evident that our \proposed can more accurately detect salient objects in various complex scenes. Even in cases where one modality fails locally ($2^{nd}_{}$-$4^{th}_{}$, $7^{th}_{}$, and $9^{th}_{}$ rows) or both modalities provide poor-quality information ($2^{nd}_{}$, $5^{th}_{}$, and $7^{th}_{}$ rows), \proposed can mitigate noise and extraneous contextual cues, enabling precise delineation of salient objects from the background. This is because our proposed MPA reliably captures long-range dependencies and effectively utilizes bimodal complementary information. Moreover, in fine-grained object detection, existing models, including UniTR \cite{UniTR}, often incur losses in object depiction. In contrast, \proposed excels in capturing intricate object details, such as the legs of a person ($2^{th}_{}$ row), a person on a motorcycle ($7^{th}_{}$ row), and the frame of two chairs ($10^{th}_{}$ row). This superior performance can be attributed to the collaboration of our proposed \proposed's components, namely FEB, FRCAB, and CFL. FEB provides reliable edge details through the depth frequency decomposition of low-level features. FRCAB enhances the model's robustness, improving its performance in various scenes. More critically, CFL effectively addresses the problem of lost visual signals and hard frequencies unattended in the frequency domain.

\begin{table*}[!htp]
  \centering
  \fontsize{8}{10}\selectfont
  \renewcommand{\arraystretch}{1}
  \renewcommand{\tabcolsep}{0.4mm}
  \scriptsize
  \caption{Ablation Study of Each Component in Our \proposed. The Best Results are Labeled \textcolor{red}{\textbf{Red}} and the Second Best are Labeled \textcolor{blue}{Blue}. FPS Was Tested on an NVIDIA RTX4060 Laptop GPU.}
\label{tab:Ablation}
  \scalebox{0.88}{
  \begin{tabular}{c|c|ccccc|ccccc|ccccc|ccccc|c|c|c}
  \hline\toprule
   \multirow{2}{*}{\centering Type} & \multirow{2}{*}{\centering Settings}  & \multicolumn{5}{c|}{\centering VT821} & \multicolumn{5}{c|}{\centering VT1000} & \multicolumn{5}{c|}{\centering VT5000} &\multicolumn{5}{c|}{\centering VI-RGBT1500}& \multirow{2}{*}{\centering Params(M)} & \multirow{2}{*}{\centering FLOPs(G)} & \multirow{2}{*}{\centering FPS}\\
   &\multicolumn{1}{c|}{} & $Em\uparrow$ & $Sm\uparrow$ & $wF_\beta \uparrow$ & $F_\beta\uparrow$ & $\mathcal{M}\downarrow$ & $Em\uparrow$ & $Sm\uparrow$ & $wF_\beta \uparrow$ & $F_\beta\uparrow$ & $\mathcal{M}\downarrow$ & $Em\uparrow$ & $Sm\uparrow$ & $wF_\beta \uparrow$ & $F_\beta\uparrow$ & $\mathcal{M}\downarrow$ & $Em\uparrow$ & $Sm\uparrow$ & $wF_\beta \uparrow$ & $F_\beta\uparrow$ & $\mathcal{M}\downarrow$ & \multicolumn{1}{c|}{} & \multicolumn{1}{c|}{} & \multicolumn{1}{c}{}\\
\midrule
    \multirow{2}{*}{Modality} & w/o Thermal &0.928 &0.905 &0.841 &0.856 &0.029 &0.950 &0.942 &0.904 &0.890 &0.021 &0.943 &0.908 &0.851 &0.868 &0.029 &0.945 &0.894 &0.844 &0.874 &0.034 &86.7 &101.2 & 24.9\\
    & w/o RGB &0.900 &0.855 &0.768 &0.786 &0.041 &0.940 &0.916 &0.875 &0.863 &0.026 &0.930 &0.885 &0.819 &0.835 &0.034 &0.901 &0.821 &0.750 &0.791 &0.058 &86.7 &101.2 & 24.9\\
\midrule
   \multirow{3}{*}{Backbone}
    & w/ GFNet-H-B & 0.898 & 0.865 & 0.749 & 0.773 & 0.044 & 0.920 & 0.887 & 0.856 & 0.860 & 0.034 & 0.893 & 0.842 & 0.752 & 0.790 & 0.042 & 0.880 & 0.804 & 0.711 & 0.777 & 0.060 & 136.5 & 129.1 & 19.4\\	
    & w/ AFNO & 0.912 & 0.879 & 0.780 & 0.806 & 0.038 & 0.935 & 0.916 & 0.878 & 0.879 & 0.024 & 0.917 & 0.874 & 0.805 & 0.828 & 0.036 &  0.918 & 0.856 & 0.783 & 0.819 & 0.047 & 132.7 & 138.0 & 7.6\\				
    & w/ DFFormer-M & 0.928 & \textcolor{blue}{0.912} & 0.861 & 0.858 & 0.026 & \textcolor{blue}{0.957} & 0.946 & \textcolor{blue}{0.925} & 0.900 & \textcolor{red}{\textbf{0.014}} & \textcolor{red}{\textbf{0.951}} & \textcolor{red}{\textbf{0.922}} & \textcolor{red}{\textbf{0.879}} & \textcolor{blue}{0.883} & \textcolor{red}{\textbf{0.023}} & \textcolor{red}{\textbf{0.956}} & \textcolor{blue}{0.915} & \textcolor{red}{\textbf{0.887}} & \textcolor{red}{\textbf{0.896}} & \textcolor{red}{\textbf{0.026}} & 151.3 & 138.0 & 11.7\\
\midrule
    \multirow{7}{*}{Module} & w/o MPA & 0.926 & 0.908 & 0.857 & 0.857 &	\textcolor{blue}{0.025} & 0.954 & 0.944 &	0.921 &	0.898 & 0.016 & 0.944 & 0.911 & 0.865 & 0.874 & 0.026 & 0.949 & 0.908 & 0.874 & 0.879 & \textcolor{blue}{0.027} & 144.2 & 137.9 & 18.2\\
    & w/ MHCA & 0.915 & 0.902 & 0.839 & 0.847 & 0.027 & 0.948 & 0.941 & 0.913 & 0.890 & 0.017 & 0.941 & 0.906 & 0.855 & 0.865 & 0.027 & 0.931 & 0.889 & 0.838 & 0.853 & 0.036 &146.3 & 138.8 & 16.7\\
    & w/o FRCAB & 0.928 & 0.904 & 0.846 & 0.849 & 0.027 & 0.952 & 0.942 & 0.917 & 0.897 & 0.016 & 0.946 & 0.910 & 0.857 & 0.870 & 0.027 & 0.950 & 0.910 & 0.870 & 0.884 & 0.029 & 127.1 & 77.4 & 17.5\\
    & w/ RCAB & 0.931 & 0.910 &	0.858 &	0.864 &	0.026 &	0.956 & \textcolor{blue}{0.947} & \textcolor{blue}{0.925} & \textcolor{blue}{0.906} & \textcolor{blue}{0.015} & \textcolor{blue}{0.948} & 0.918 & \textcolor{blue}{0.876} & 0.880 & \textcolor{blue}{0.024} & 0.953 & 0.911 & 0.881 & 0.885 & \textcolor{blue}{0.027} & 148.1 & 139.5 & 17.3\\
    & w/o FEB & 0.928 & 0.911 & 0.855 & 0.850 & 0.026 & 0.956 & \textcolor{red}{\textbf{0.948}} & 0.919 & 0.892 & \textcolor{blue}{0.015} & 0.947 & 0.919 & 0.869 & 0.872 & 0.025 & 0.949 & 0.913 & 0.872 & 0.878 & 0.029 & 148.6 & 135.6 & 17.8\\
    & w/o CFL & \textcolor{blue}{0.933} & 0.911 & \textcolor{blue}{0.864} & \textcolor{blue}{0.865} & \textcolor{red}{\textbf{0.023}} & 0.956 & 0.945 & \textcolor{blue}{0.925} & \textcolor{red}{\textbf{0.909}} & \textcolor{blue}{0.015} & 0.947 & 0.916 & 0.874 & 0.881 & \textcolor{blue}{0.024} & \textcolor{blue}{0.954} & 0.908 & 0.879 & 0.887 & 0.028 & 149.6 & 139.5 & 17.1\\
\midrule
    & Full model & \textcolor{red}{\textbf{0.939}} & \textcolor{red}{\textbf{0.917}} & \textcolor{red}{\textbf{0.873}} & \textcolor{red}{\textbf{0.872}} & \textcolor{red}{\textbf{0.023}} & \textcolor{red}{\textbf{0.959}} & \textcolor{red}{\textbf{0.948}} & \textcolor{red}{\textbf{0.928}} & 0.903 & \textcolor{red}{\textbf{0.014}} & \textcolor{red}{\textbf{0.951}} & \textcolor{blue}{0.920} & \textcolor{red}{\textbf{0.879}} & \textcolor{red}{\textbf{0.884}} & \textcolor{red}{\textbf{0.023}} & \textcolor{red}{\textbf{0.956}} & \textcolor{red}{\textbf{0.916}} & \textcolor{blue}{0.885} & \textcolor{blue}{0.893} & \textcolor{red}{\textbf{0.026}} & 149.6 & 139.5 & 17.1\\
    \bottomrule
    \hline
  \end{tabular}}
\end{table*}

\begin{figure*}[!htp]
	\centering \includegraphics[width=0.9\textwidth]{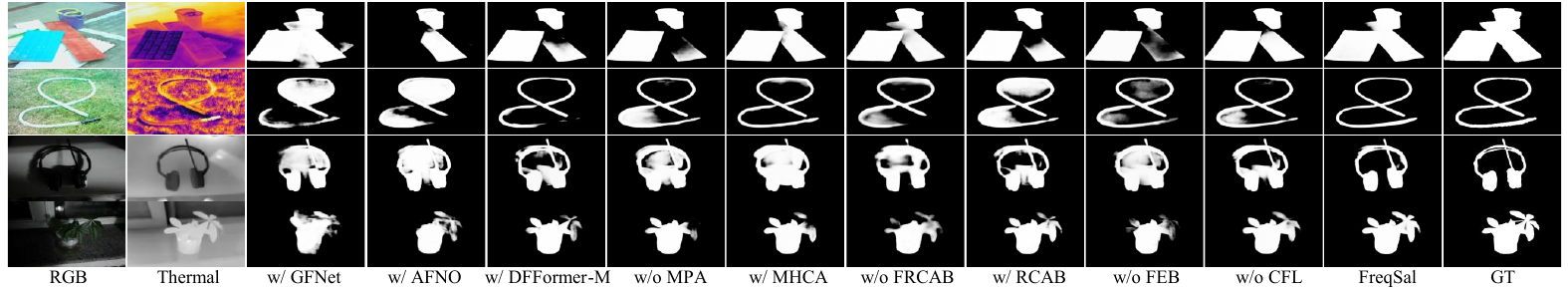}
	\caption{Visualization results of ablation experiments. GT is ground-truth.}
\label{fig:Ablation}
\end{figure*}

\subsection{Ablation Study} \label{sec:Ablation Study}
\begin{figure*}[!htb]
	\centering \includegraphics[width=0.90\textwidth]{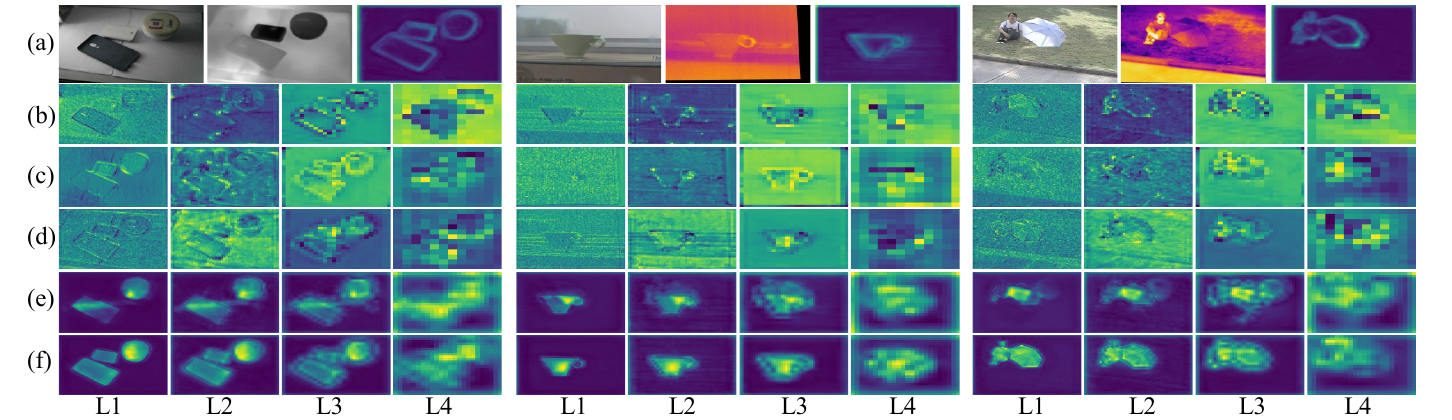}
	\caption{Visualization of the feature maps pre- and post-processed by each component. L1 to L4 indicate the layers from low to high. (a) shows RGB images, thermal images, and the feature map after FEB. (b) and (c) stand for RGB and thermal feature maps generated by the encoder. (d) represents the fused feature map obtained by MPA. (e) presents the feature maps after RCAB. (f) is the feature maps after FRCAB.}
\label{fig:attention}
\end{figure*}

To comprehensively assess the effectiveness of each component in our proposed \proposed, we conducted ablation studies on the VT821, VT000, VT5000, and VI-RGBT1500 datasets. We compared models with (w/) and without (w/o) these components, and replaced them with classical methods that operate similarly. The quantitative and qualitative evaluations of ablation studies are shown in Table \ref{tab:Ablation}, Fig. \ref{fig:Ablation}, and Fig. \ref{fig:attention}.

\emph{1) Effectiveness of auxiliary modality:}
To evaluate the contribution of each auxiliary modality, we conducted experiments by removing the thermal and RGB modalities separately. 
As shown in the $1^{st}_{}$ and $2^{nd}_{}$ rows of Table \ref{tab:Ablation}, the model's performance declined in both cases. 
In particular, excluding the RGB modality led to average decreases of 3.5\%, 6.1\%, 9.9\%, and 7.8\% in $Em$, $Sm$, ${wF_\beta}$, and ${F_\beta}$, respectively. 
These results confirm that both modalities are crucial for achieving accurate detection.

\emph{2) Effectiveness of backbone network:} To validate the effectiveness of CDFFormer-M \cite{tatsunami2024fft} as the backbone of \proposed, we compared it with three other FFT-based backbones, namely GFNet-H-B \cite{rao2021global}, AFNO \cite{guibas2021adaptive}, and DFFormer-M \cite{tatsunami2024fft}. Notably, AFNO had the same dimension and depth as CDFFormer-M. As shown in the $3^{th}_{}$-$4^{th}_{}$ rows of Table \ref{tab:Ablation}, the model with GFNet-H-B and AFNO yielded unsatisfactory results, with inferior performance across all four datasets compared to the model with CDFFormer-M. 
This trend was further highlighted in Fig. \ref{fig:Ablation}, where these models struggled to accurately detect salient objects. Meanwhile, the model with DFFormer-M demonstrated uneven performance on the four datasets. Although it exhibited a slight advantage on the VI-RGBT1500 dataset ($5^{th}_{}$ row of Table \ref{tab:Ablation}), it substantially lagged behind the model with CDFFormer-M on the VT821 dataset.

In addition, we attempted to apply spatial domain-based backbones to \proposed, such as VGG \cite{simonyan2014very}, ResNet \cite{he2016deep}, and Swin Transformer \cite{liu2021swin}. 
However, successive Inverse Fourier Transforms of MPA in the feature space of the spatial domain backbone lead to a quadratic amplification of the gradient, resulting in instability during training (loss reported as ``NaN'').

\emph{3) Effectiveness of MPA (Sec. \ref{sec:MPA}):} First, we removed MPA and employed element-wise addition to fuse the bimodal features, labeled ``w/o MPA" in Table \ref{tab:Ablation} and Fig. \ref{fig:Ablation}. 
The performance of the model lacking MPA declined obviously on the four datasets due to its inability to extract complementary information. In Fig. \ref{fig:Ablation}, when the salient region in one or both modalities shows a different appearance, the model without MPA encounters considerable interference, impeding accurate detection of complete objects.

To further validate the effectiveness of the 
MPF, the core of MPA, we replaced MPF with the widely used multi-head cross-attention (MHCA) \cite{VST,10015667}. Due to the considerable complexity and memory demands of MHCA, we applied it only to the last three low-resolution layers, while the highest resolution layer still employed element-wise addition. As shown in the $7^{th}_{}$ row of Table \ref{tab:Ablation} and Fig. \ref{fig:Ablation}, the model with MHCA did not perform as well as expected, and even had less gain than the model with element-wise addition.
The experiments highlight the suboptimal performance of MHCA compared to MPA in high-resolution feature fusion and demonstrate the incompatibility of spatial-domain operations with frequency-domain models.

Finally, we visualized in Fig. \ref{fig:attention} the feature changes after processing by MPA. Even in instances of localized failure within the RGB (\textit{e.g.}, case 1) or thermal modalities (\textit{e.g.}, cases 2 and 3), MPA demonstrates remarkable resilience, successfully fusing both modalities without succumbing to noise interference. The outcomes of the above experiments collectively reinforce the high efficacy of MPA.

\emph{4) Effectiveness of FRCAB (Sec. \ref{sec:FRCAB}):} To verify the effectiveness of the proposed FRCAB, we first replaced it with $DNRU(\cdot)$ in Eq. (\ref{eq:21}), labeled “w/o FRCAB” in Table \ref{tab:Ablation} and Fig. \ref{fig:Ablation}. The results indicate a significant decrease in metrics $Em$, $Sm$, ${wF_\beta}$, ${F_\beta}$, and $\mathcal{M}$, with average reductions of 0.8$\%$, 0.9$\%$, 2.1$\%$, 1.5$\%$, and 23.2$\%$, respectively. 
The visualization results confirmed this conclusion, revealing that the model would be more susceptible to cluttered backgrounds, lighting variations, and thermal imbalances without FRCAB.

Subsequently, we replaced FRCAB with the 
RCAB \cite{zhang2018image}, labeled ``w/ RCAB" in Table \ref{tab:Ablation} and Fig. \ref{fig:Ablation}. Comparing the $9^{th}_{}$ and last rows of Table \ref{tab:Ablation}, we found that while the RCAB-equipped model performs well, it consistently falls short of the FRCAB-equipped model across all metrics. 
Additionally, as shown in Fig. \ref{fig:attention}, the features processed by RCAB exhibit attention drift and edge blurring, whereas those processed by FRCAB effectively decode the fused features after MPA and focus more accurately on salient regions. By modeling channel-wise dependencies, FRCAB enables the model to prioritize high-frequency information and gradually generate high-resolution features.
This clearly demonstrates that FRCAB can stably improve the model's robustness and performance compared to RCAB in our \proposed. 


\emph{5) Effectiveness of FEB (Sec. \ref{sec:FEB}):} To validate the gains made by the proposed FEB, we replaced its core component 
EFEB with $DNR(\cdot)$ in Eq. (\ref{eq:21}), while retaining FRCAB, and presented the results in the $10^{th}_{}$ row of Table \ref{tab:Ablation}. The comparison reveals that without EFEB, all metrics of the model exhibit a decrease to varying degrees. Furthermore, the visualization results depicted in Fig. \ref{fig:Ablation} fail to portray complete object boundaries. In contrast, as illustrated in Fig. \ref{fig:attention}(a), the complete FEB consistently provides accurate edge features, regardless of the complexity of the scene. This ability to capture precise edge information offers a powerful boost to the model.

\begin{figure}[!htp]
	\centering \includegraphics[width=0.48\textwidth]{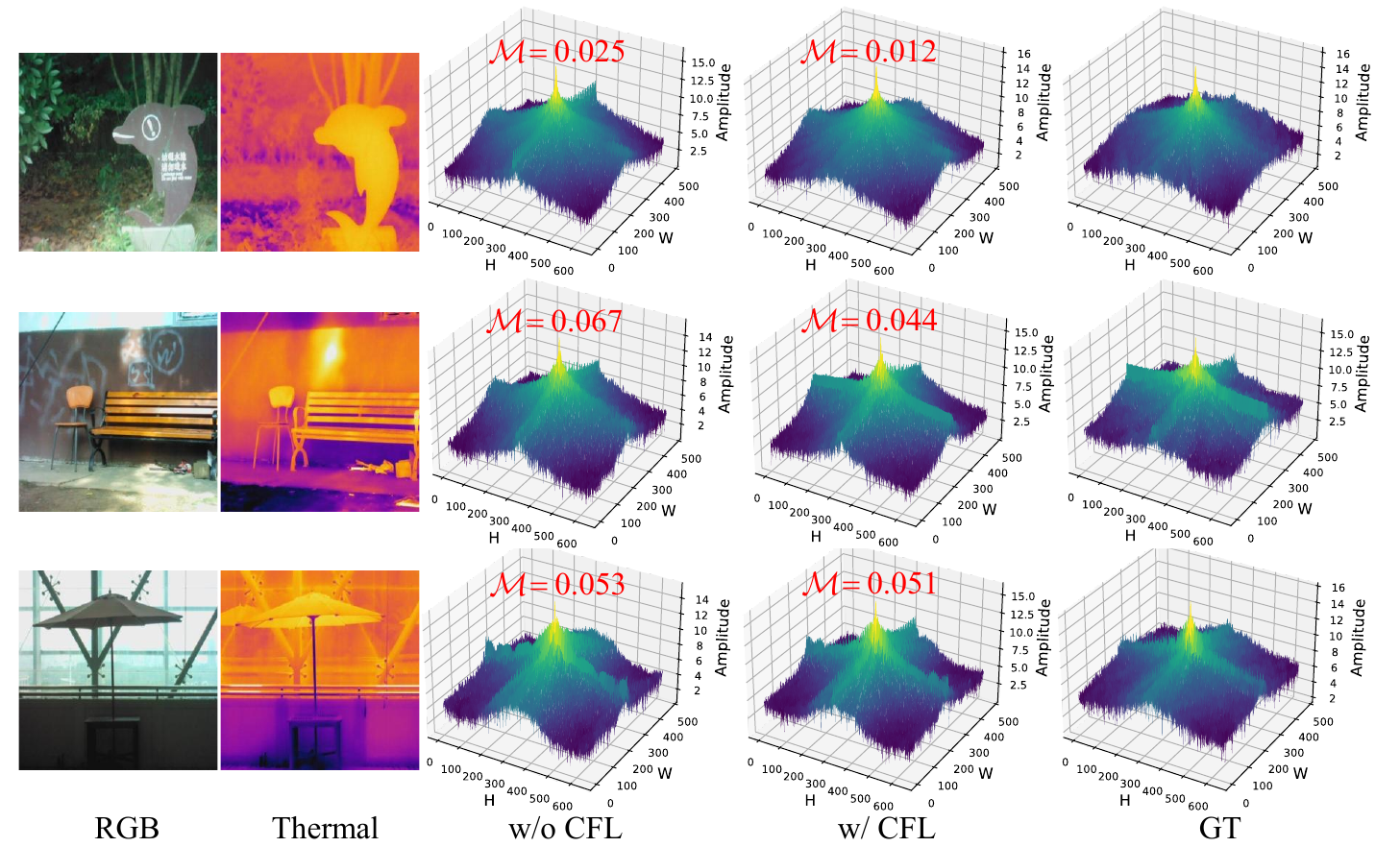}
	\caption{Frequency spectrum visualization of results from the model training with (w/) and without (w/o) CFL. GT stands for ground-truth.}
\label{fig:Fre3D}
\end{figure}
\begin{figure}[!htp]
	\centering \includegraphics[width=0.48\textwidth]{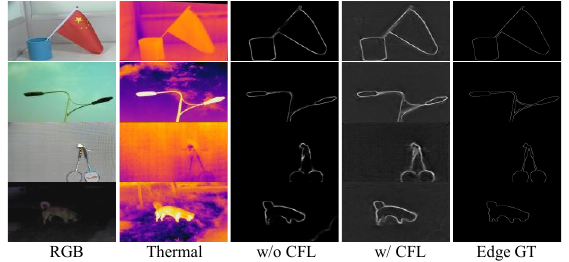}
	\caption{Visual comparison of edge maps between the model (w/) and without (w/o) CFL. Edge GT stands for the ground-truth of edge.}
\label{fig:comp_edge}
\end{figure}

\emph{6) Effectiveness of CFL (Sec. \ref{sec:bidomain}):} To evaluate the contribution of CFL, we removed it and kept only BCE to supervise edge learning, denoted as ``w/o CFL" in Table \ref{tab:Ablation}, Fig. \ref{fig:Ablation} and \ref{fig:comp_edge}. The results show a degradation in performance on the four datasets after losing the supervision of CFL. 

Moreover, we displayed the 3D frequency spectra of the results obtained with both settings in Fig. \ref{fig:Fre3D}. It can be seen that 3D spectrograms constructed without CFL supervision exhibit oscillations and frequency distributions that deviate further from the ground-truth. Conversely, the presence of CFL optimizes the loss of frequency reconstruction and brings the frequency statistics closer to ground-truth. 

Furthermore, we presented a visual comparison of the edge maps for the two settings in Fig. \ref{fig:comp_edge}. Even when both modalities provide clear objects (\textit{i.e.}, cases in the $1^{st}$-$2^{nd}$ rows), 
the model trained only with BCE still focuses more on edge locations with high priority, making it difficult to detect complete object edges. 
\begin{table}[!htp]
  \centering
  \fontsize{8}{10}\selectfont
  \renewcommand{\arraystretch}{1}
  \renewcommand{\tabcolsep}{0.8mm}
  \scriptsize
  \caption{Quantitative Comparison of the Proposed FreqSal with High-resolution Inputs. The Best Results are Labeled \textcolor{red}{\textbf{Red}}. FPS Was Tested on an NVIDIA RTX4060 Laptop GPU.}
\label{tab:input_size}
  \scalebox{0.85}{
  \begin{tabular}{c|cc|cc|cc|cc|c|c|c}
  \hline\toprule
    \multirow{2}{*}{\centering Input Size}  & \multicolumn{2}{c|}{\centering VT821} & \multicolumn{2}{c|}{\centering VT1000} & \multicolumn{2}{c|}{\centering VT5000} &\multicolumn{2}{c|}{\centering VI-RGBT1500} & \multirow{2}{*}{\centering Param(M)} & \multirow{2}{*}{\centering FLOPs(G)} & \multirow{2}{*}{\centering FPS}\\
   \multicolumn{1}{c|}{} & $F_\beta\uparrow$ & $\mathcal{M}\downarrow$ & $F_\beta\uparrow$ & $\mathcal{M}\downarrow$ & $F_\beta\uparrow$ & $\mathcal{M}\downarrow$ & $F_\beta\uparrow$ & $\mathcal{M}\downarrow$ & \multicolumn{1}{c|}{} & \multicolumn{1}{c|}{} & \multicolumn{1}{c}{}\\
\midrule
    512$^2$ & 0.869 & 0.024 & \textcolor{red}{\textbf{0.903}} & 0.015 & \textcolor{red}{\textbf{0.885}} & \textcolor{red}{\textbf{0.023}} & 0.891 & \textcolor{red}{\textbf{0.026}} & 149.6 & 248.0 & 12.6\\	
    768$^2$ & 0.843 &  0.031 &  0.882 & 0.022 & 0.860 & 0.029 & 0.876 & 0.033 & 149.6 & 558.0 & 4.8\\
\midrule
    384$^2$ & \textcolor{red}{\textbf{0.872}} & \textcolor{red}{\textbf{0.023}} & \textcolor{red}{\textbf{0.903}} & \textcolor{red}{\textbf{0.014}} & 0.884 & \textcolor{red}{\textbf{0.023}} & \textcolor{red}{\textbf{0.893}} & \textcolor{red}{\textbf{0.026}}& 149.6 & 139.5 & 17.1\\
    \bottomrule
    \hline
  \end{tabular}}
\end{table}
In contrast, the model supervised with additional CFL not only demonstrates trustworthy performance in typical scenarios but also presents explicit object edges by learning difficult frequencies, 
even when object edges are difficult to distinguish from complex scenarios (\textit{i.e.} $3^{rd}$-$4^{th}$ rows).
In conclusion, these findings strongly demonstrate the effectiveness of CFL in learning accurate edge maps and its essential role in \proposed.

\emph{7) Increase input resolution:} 
To verify the effectiveness of \proposed in handling high-resolution bimodal images, we increased the input resolution to $512^{2}$ and $768^{2}$ (trained on an NVIDIA RTX3090 GPU with 24GB of memory).
A comparison in Table \ref{tab:input_size} showed that the $512^{2}$ resolution model has comparable performance to the $384^{2}$ resolution model, while the $768^{2}$ resolution model experiences a noticeable performance decline.
This performance degradation is likely due to the lower input resolution of the pre-trained encoder and the smaller batch size adopted during training. Significantly, this experiment represents the first exploration of high-resolution RGB-T SOD, whereas prior models typically operate within input resolution ranging from $224^{2}$ to $384^{2}$.

\subsection{Generalization Experiments}
\emph{1) RGB-D-T SOD:}
We compared our \proposed with thirteen existing models, which consist of (1) ten bimodal models, namely CSRNet \cite{9505635}, CGFNet \cite{wang2021cgfnet}, DCNet \cite{tu2022weakly}, SwinNet \cite{9611276}, LSNet \cite{10042233}, HRTNet \cite{9869666}, PICRNet \cite{PICRNet}, LAFB \cite{wang2024learning}, MAGNet \cite{MAGNet}, and UniTR \cite{UniTR}, (2) three trimodal models, namely HWSI \cite{HWSI}, MFFNet \cite{MFFNet}, and QSFNet \cite{QSFNet}. To ensure a fair comparison, all saliency maps used for comparison were obtained from the corresponding models' homepages or generated using open-source code.

\begin{table}[!htp]
  \centering
  \fontsize{8}{10}\selectfont
  \renewcommand{\arraystretch}{1}
  \renewcommand{\tabcolsep}{0.5mm}
  \scriptsize
  \caption{Quantitative Comparison of 
  Our \proposed and Thirteen State-of-the-Art Models on the VDT-2048 dataset. VT, VD, and VDT Denote Training and Testing on RGB-T, RGB-D, and RGB-D-T Data, Respectively. The Best Results are Labeled \textcolor{red}{\textbf{Red}} and the Second Best Results are Labeled \textcolor{blue}{Blue}.}
\label{tab:vdt}
  \scalebox{1}{
  \begin{tabular}{c|c|c|ccccc}
  \hline\toprule
   \multirow{2}{*}{\centering Model} & \multirow{2}{*}{\centering Tpye} & \multirow{2}{*}{\centering Backbone}  & \multicolumn{5}{c}{\centering VDT2048}\\
   &\multicolumn{1}{c|}{} &\multicolumn{1}{c|}{} & $Em\uparrow$ & $Sm\uparrow$ & $wF_\beta \uparrow$ & $F_\beta\uparrow$ & $\mathcal{M}\downarrow$\\
\midrule
    \multicolumn{8}{c}{\centering Bimodal Model}\\
\midrule
   CSRNet$_{21}$ \cite{9505635} & VT & ESPNetv2 & 0.9494 & 0.8821 & 0.8099 & 0.7888 &	0.0050\\
   CGFNet$_{21}$ \cite{wang2021cgfnet} & VT & VGG-16 & 0.9319 & 0.9166 & 0.8425 & 0.7822 & 0.0033\\
   DCNet$_{22}$ \cite{tu2022weakly} & VT & VGG-16 & 0.9658	& 0.8787 & 0.8227 & 0.8521 & 0.0038\\
   SwinNet$_{22}$ \cite{9611276} & VT & Swin-B & 0.9444 & \textcolor{blue}{0.9370} & 0.8841 & 0.8090 & 0.0026\\
   SwinNet$_{22}$ \cite{9611276} & VD & Swin-B & 0.8978 & 0.9198 & 0.8344 & 0.7321 & 0.0037\\
   LSNet$_{23}$ \cite{10042233} & VT & MobileNetv2 & 0.9327 & 0.8867 &0.8005 & 0.7607	& 0.0044\\
   HRTNet$_{23}$ \cite{9869666} & VT & HRFormer & 0.9680 &	0.9281&	0.8840&	0.8446 & 0.0026\\
   HRTNet$_{23}$ \cite{9869666} & VD & HRFormer & 0.9617 & 0.9144 &	0.8603 & 0.8227 & 0.0031\\
   PICRNet$_{23}$ \cite{PICRNet} & VD & Swin-T & 0.9574 & 0.8941 & 0.8267 & 0.7998 & 0.0040\\
   LAFB$_{24}$ \cite{wang2024learning} & VT & Res2Net-50 & 0.9736 & 0.9325 & \textcolor{blue}{0.8943} & 0.8554 & \textcolor{blue}{0.0025}\\
   LAFB$_{24}$ \cite{wang2024learning} & VD & Res2Net-50 & 0.9676 & 0.9160 & 0.8668 & 0.8356 & 0.0031\\
   MAGNet$_{24}$ \cite{MAGNet} & VT & SMT & 0.9758 & 0.9266 & 0.8848 & 0.8563 & 0.0028\\
   MAGNet$_{24}$ \cite{MAGNet} & VD & SMT & 0.9647 & 0.9092 & 0.8533 & 0.8222 & 0.0034\\
   UniTR$_{24}$ \cite{UniTR} & VT & Swin-B & \textcolor{blue}{0.9805} & 0.9084 & 0.8704 & \textcolor{blue}{0.8789} & 0.0028\\
   UniTR$_{24}$ \cite{UniTR} & VD & Swin-B & 0.9767 & 0.8930 & 0.8420 & 0.8546 & 0.0034\\
\midrule
   \proposed (Ours) & VT & CDFFormer-M & \textcolor{red}{\textbf{0.9862}} & \textcolor{red}{\textbf{0.9384}} & \textcolor{red}{\textbf{0.8954}} & \textcolor{red}{\textbf{0.8809}} & \textcolor{red}{\textbf{0.0024}}\\
\midrule
    \multicolumn{8}{c}{\centering Trimodal Model}\\
\midrule
   HWSI$_{22}$ \cite{HWSI} & VDT & VGG-16 & 0.9815 & 0.9318 & 0.8974 & 0.8718 & 0.0026\\
   MFFNet$_{24}$ \cite{MFFNet} & VDT & VGG-16 & 0.9831 & \textcolor{blue}{0.9394} & \textcolor{blue}{0.9017} & 0.8758 & 0.0025\\
   QSFNet$_{24}$ \cite{QSFNet} & VDT & Swin-B & \textcolor{red}{\textbf{0.9868}} & \textcolor{red}{\textbf{0.9426}} & \textcolor{red}{\textbf{0.9145}} & \textcolor{red}{\textbf{0.8902}} & \textcolor{red}{\textbf{0.0021}}\\
\midrule
   \proposed (Ours) & VT & CDFFormer-M & \textcolor{blue}{0.9862} & 0.9384 & 0.8954 & \textcolor{blue}{0.8809} & \textcolor{blue}{0.0024}\\
    \bottomrule
    \hline
  \end{tabular}}
\end{table}

\begin{figure}[!htp]
	\centering \includegraphics[width=0.49\textwidth]{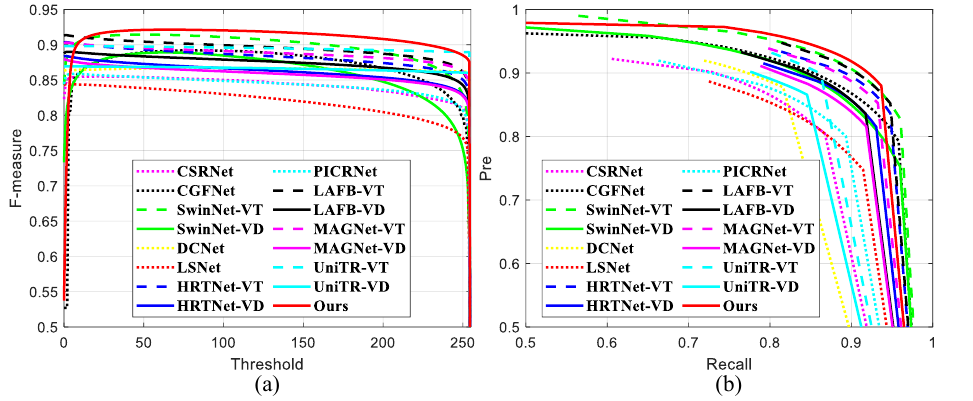}
	\caption{(a) F-measure-threshold (FT) and (b) Precision-recall (PR) curves of our \proposed and existing ten bimodal SOD models on the VDT-2048 dataset.}
\label{fig:FTPR_VDT}
\end{figure}

Table \ref{tab:vdt} and Fig. \ref{fig:FTPR_VDT} present the quantitative comparison of the proposed \proposed with thirteen SOTA models on the VDT-2048 dataset.
In summary, our \proposed surpassed ten existing bimodal models across all six metrics and achieved performance comparable to trimodal models while using only two modalities.
Specifically, compared to the leading bimodal models on RGB-T data, \proposed surpassed UniTR \cite{UniTR} by 0.6\% and 0.2\% in $Em$ and $F_\beta$, respectively, and exceeded LAFB \cite{wang2024learning} by 0.1\% and 4.0\% in $wF_\beta$ and $\mathcal{M}$, respectively. 
Compared to trimodal models, \proposed trained on RGB-T data solidly outperformed HWSI \cite{HWSI} and led MFFNet \cite{MFFNet} in $Em$, $F_\beta$, and $\mathcal{M}$. 
Although it lagged behind QSFNet \cite{QSFNet}, our \proposed remained competitive with trimodal models.
These findings highlight the effectiveness and strong generalization ability of our \proposed on the VDT-2048 dataset.

\emph{2) RGB-D SOD:}
We conducted a comprehensive comparison with sixteen existing RGB-D SOD models, namely
LSNet \cite{10042233}, PopNet \cite{PopNet}, C$^2$DFNet \cite{C2DFNet}, HiDANet \cite{HiDANet}, RD3D+ \cite{RD3D+}, FasterSal \cite{Fastersal}, LAFB \cite{wang2024learning}, VST \cite{VST}, SwinNet \cite{9611276}, CAVER \cite{10015667}, PICRNet \cite{PICRNet}, MAGNet \cite{MAGNet}, HFMDNet \cite{HFMDNet}, EM-Trans \cite{EM-Trans}, VST++ \cite{liu2024vst++}, and UniTR \cite{UniTR}.
To ensure a fair comparison, all saliency maps were either obtained from the authors of the corresponding models or generated using their publicly available code.

\begin{figure*}[ht]
	\centering \includegraphics[width=1\textwidth]{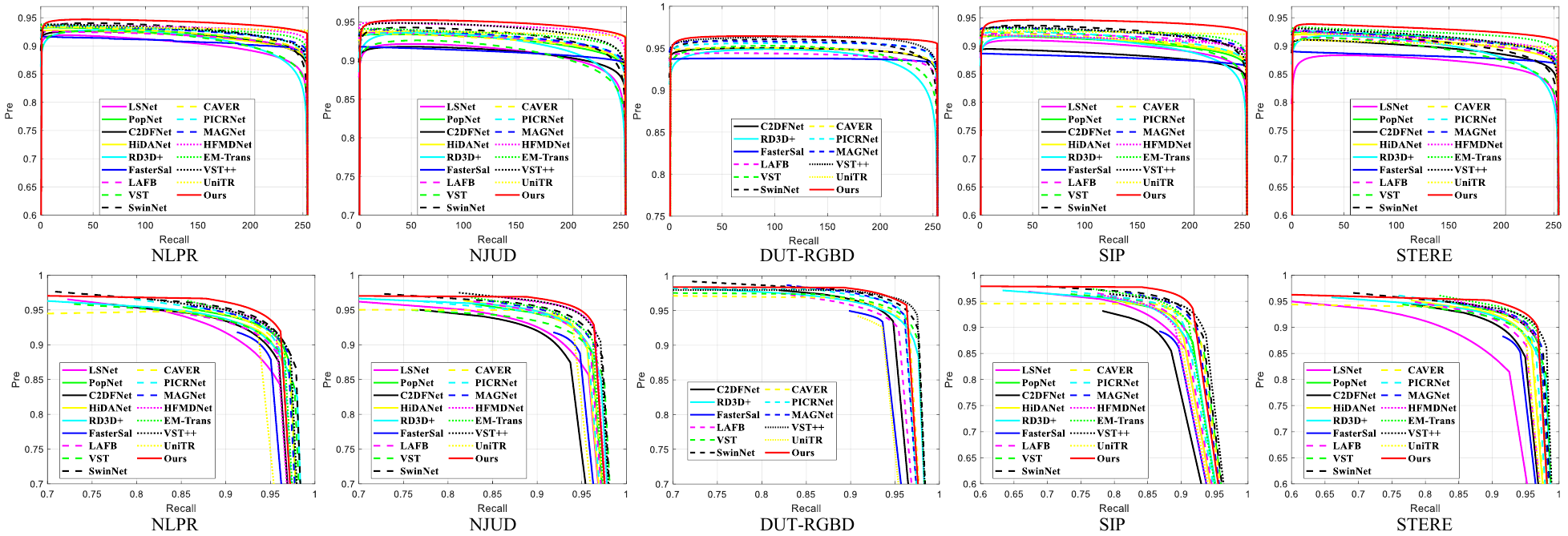}
	\caption{F-measure curves and precision-recall (PR) curves of the proposed \proposed and sixteen existing RGB-D SOD models on the NLPR, NJUD, DUT-RGBD, SIP, and STERE datasets.}
\label{fig:FTPR_RGBD}
\end{figure*}

Table \ref{tab:RGBDcomparison} and Fig. \ref{fig:FTPR_RGBD} present the comparison of the proposed \proposed with sixteen RGB-D SOD models on the NLPR, NJUD, DUT-RGBD, SIP, and STERE datasets using six evaluation metrics: $Em$, $Sm$, $wF_\beta$, $F_\beta$, $\mathcal{M}$, and PR curve. 
Overall, our \proposed surpassed sixteen existing RGB-D models across all six evaluation metrics on the five datasets. 
Specifically, \proposed outperformed the pervious best-performing model, VST++ \cite{liu2024vst++}, achieving average improvements of 0.9\%, 0.4\%, 1.1\%, 0.5\%, 7.0\% in $Em$, $Sm$, $wF_\beta$, $F_\beta$, and $\mathcal{M}$, respectively. 
Compared to HFMDNet \cite{HFMDNet}, \proposed shows a substantial performance advantage on the SIP and STERE datasets, while maintaining comparable results on the NLPR and NJUD datasets.
This strong performance highlights the generalization ability of FreqSal's deep Fourier-embedded strategy, which proves equally effective in addressing the RGB-D SOD task. 

\begin{table*}[!htp]
  \centering
  \fontsize{8}{10}\selectfont
  \renewcommand{\arraystretch}{1.1}
  \renewcommand{\tabcolsep}{0.35mm}
  \scriptsize
  \caption{Quantitative Comparison of 
  Our \proposed and Sixteen State-of-the-Art RGB-D SOD Models on the NLPR, NJUD, DUT-RGBD, SIP, and STERE datasets. The Best Results are Labeled \textcolor{red}{\textbf{Red}} and the Second Best Results are Labeled \textcolor{blue}{Blue}.}
\label{tab:RGBDcomparison}
  \scalebox{0.9}{
  \begin{tabular}{c|ccccc|ccccc|ccccc|ccccc|ccccc}
  \hline\toprule
   \multirow{2}{*}{\centering Model} & \multicolumn{5}{c|}{\centering NLPR} & \multicolumn{5}{c|}{\centering NJUD} & \multicolumn{5}{c|}{\centering DUT-RGBD} & \multicolumn{5}{c|}{\centering SIP} & \multicolumn{5}{c}{\centering STERE}\\
   & $Em\uparrow$ & $Sm\uparrow$ & $wF_{\beta} \uparrow$ & $F_{\beta}\uparrow$ & $\mathcal{M}\downarrow$ &$Em\uparrow$ & $Sm\uparrow$ &$wF_{\beta} \uparrow$ & $F_{\beta}\uparrow$ & $\mathcal{M}\downarrow$ & $Em\uparrow$ & $Sm\uparrow$ & $wF_{\beta} \uparrow$ & $F_{\beta}\uparrow$ & $\mathcal{M}\downarrow$ & $Em\uparrow$ & $Sm\uparrow$ & $wF_{\beta} \uparrow$ & $F_{\beta}\uparrow$ & $\mathcal{M}\downarrow$ & $Em\uparrow$ & $Sm\uparrow$ & $wF_{\beta} \uparrow$ & $F_{\beta}\uparrow$ & $\mathcal{M}\downarrow$\\
\midrule
    \multicolumn{26}{c}{\centering CNN-based Model}\\
\midrule
    LSNet$_{23}$ \cite{10042233} & 0.955 & 0.919 & 0.873 & 0.881 & 0.025 & 0.922 & 0.911 & 0.879 & 0.899 & 0.038 & - & - & - & - & - & 0.927 & 0.886 & 0.848 & 0.882 & 0.049 & 0.913 & 0.871 & 0.817 & 0.854 & 0.054\\
    PopNet$_{23}$ \cite{PopNet} & 0.961 & 0.936 & 0.904 & 0.903 & 0.020 & 0.929 & 0.925 & 0.904 & 0.916 & 0.031 & - & - & - & - & - & 0.929 & 0.894 & 0.867 & 0.890 & 0.043 & 0.931 & 0.920 & 0.889 & 0.903 & 0.033\\
    C$^2$DFNet$_{23}$ \cite{C2DFNet} & 0.958 & 0.928 & 0.890 & 0.897 & 0.021 & 0.919 & 0.908 & 0.879 & 0.898 & 0.038 & 0.958 & 0.933 & 0.919 & 0.933 & 0.025 & 0.916 & 0.871 & 0.831 & 0.865 & 0.053 & 0.926 & 0.902 & 0.863 & 0.881 & 0.038\\
    HiDANet$_{23}$ \cite{HiDANet}& 0.959 & 0.930 & 0.898 & 0.899 & 0.022 & 0.935 & 0.926 & 0.908 & 0.919 & 0.030 & - & - & - & - & - & 0.925 & 0.893 & 0.864 & 0.889 & 0.044 & 0.934 & 0.912 & 0.880 & 0.894 & 0.036\\
    RD3D+$_{24}$ \cite{RD3D+} & 0.958 & 0.933 & 0.883 & 0.886 & 0.022 & 0.920 & 0.928 & 0.894 & 0.910 & 0.033 & 0.953 & 0.936 & 0.909 & 0.945 & 0.031 & 0.924 & 0.892 & 0.850 & 0.883 & 0.046 & 0.921 & 0.914 & 0.860 & 0.881 & 0.039\\
    FasterSal$_{24}$ \cite{Fastersal} & 0.955 & 0.922 & 0.890 & 0.895 & 0.023 & 0.919 & 0.909 & 0.891 & 0.902 & 0.034 & 0.952 & 0.920 & 0.909 & 0.922 & 0.031 & 0.924 & 0.872 & 0.841 & 0.867 & 0.049 & 0.918 & 0.890 & 0.856 & 0.872 & 0.040\\
    LAFB$_{24}$ \cite{wang2024learning} & 0.958 & 0.930 & 0.895 & 0.902 & 0.022 & 0.924 & 0.924 & 0.904 & 0.918 & 0.029 & 0.957 & 0.931 & 0.915 & 0.929 & 0.027 & 0.937 & 0.897 & 0.875 & 0.904 & 0.041 & 0.930 & 0.907 & 0.875 & 0.895 & 0.037\\
\midrule
    \multicolumn{26}{c}{\centering Transformer-based Model}\\
\midrule
    VST$_{21}$ \cite{VST} & 0.954 & 0.931 & 0.887 & 0.886 & 0.023 & 0.914 & 0.922 & 0.888 & 0.900 & 0.034 & 0.960 & 0.943 & 0.920 & 0.921 & 0.025 & 0.937 & 0.904 & 0.873 & 0.895 & 0.040 & 0.917 & 0.913 & 0.866 & 0.878 & 0.038\\
    SwinNet$_{22}$ \cite{9611276} & 0.967 & 0.941 & 0.908 & 0.908 & \textcolor{blue}{0.018} & 0.934 & 0.935 & 0.913 & 0.922 & 0.027 & 0.968 & 0.949 & 0.935 & 0.943 & \textcolor{blue}{0.020} & \textcolor{blue}{0.943} & 0.911 & 0.890 & 0.912 & 0.035 & 0.929 & 0.919 & 0.882 & 0.893 & 0.033\\
    CAVER$_{23}$ \cite{10015667} & 0.964 & 0.934 & 0.904 & 0.906 & 0.021 & 0.931 & 0.927 & 0.909 & 0.920 & 0.028 & 0.962 & 0.937 & 0.926 & 0.936 & 0.026 & 0.937 & 0.904 & 0.879 & 0.898 & 0.037 & 0.935 & 0.917 & 0.888 & 0.901 & 0.032\\
    PICRNet$_{23}$ \cite{PICRNet} & 0.965 & 0.935 & 0.904 & 0.905 & 0.019 & 0.932 & 0.927 & 0.907 & 0.917 & 0.029 & 0.965 & 0.944 & 0.933 & 0.944 & 0.021 & 0.933 & 0.899 & 0.875 & 0.897 & 0.040 & 0.936 & 0.920 & 0.891 & 0.903 & 0.031\\
    MAGNet$_{24}$ \cite{MAGNet} & 0.964 & 0.938 & 0.910 & 0.912 & \textcolor{blue}{0.018} & 0.928 & 0.929 & 0.912 & 0.922 & 0.027 & 0.966 & 0.944 & 0.935 & \textcolor{blue}{0.946} & 0.021 & 0.941 & 0.907 & 0.890 & 0.911 & 0.036 & 0.930 & 0.922 & 0.893 & 0.903 & 0.030\\
    HFMDNet$_{24}$ \cite{HFMDNet} & 0.968 & 0.938 & \textcolor{blue}{0.916} & 0.924 & \textcolor{red}{\textbf{0.017}} & 0.939 & 0.937 & \textcolor{blue}{0.928} &  \textcolor{red}{\textbf{0.937}} & \textcolor{red}{\textbf{0.023}} & - & - & - & - & - & 0.925 & 0.886 & 0.868 & 0.897 & 0.044 & 0.933 & 0.918 & 0.892 & 0.905 & 0.031\\
    EM-Trans$_{24}$ \cite{EM-Trans} & 0.966 & \textcolor{blue}{0.939} & 0.913 & 0.919 & \textcolor{red}{\textbf{0.017}} & 0.934 & 0.930 & 0.915 & 0.926 & 0.027 & - & - & - & - & - & 0.938 & 0.904 & 0.886 & 0.912 & 0.039 & \textcolor{blue}{0.938} & 0.925 & \textcolor{blue}{0.900} & \textcolor{red}{\textbf{0.913}} & \textcolor{blue}{0.028}\\
    VST++$_{24}$ \cite{liu2024vst++} & 0.963 & 0.938 & 0.910 & 0.912 & 0.020 & \textcolor{blue}{0.941} & \textcolor{blue}{0.940} & 0.925 & \textcolor{blue}{0.933} & \textcolor{blue}{0.024} & \textcolor{blue}{0.969} & \textcolor{red}{\textbf{0.953}} & \textcolor{blue}{0.942} & 0.943 & \textcolor{blue}{0.020} & 0.942 & \textcolor{blue}{0.912}  & \textcolor{blue}{0.893} & 0.911  & \textcolor{blue}{0.034} & 0.928 & \textcolor{blue}{0.926} & 0.896 & 0.906 & 0.030\\
    UniTR$_{24}$ \cite{UniTR} & \textcolor{blue}{0.970} & 0.930 & 0.912 & \textcolor{blue}{0.925} & \textcolor{blue}{0.018} & 0.940 & 0.925 & 0.918 & 0.931 & 0.025 & - & - & - & - & - & 0.940 & 0.896 & 0.888 & \textcolor{blue}{0.915} & 0.037 & 0.933 & 0.907 & 0.884 & 0.897 & 0.032\\
\midrule
    \proposed (Ours) & \textcolor{red}{\textbf{0.974}} & \textcolor{red}{\textbf{0.944}} & \textcolor{red}{\textbf{0.922}} & \textcolor{red}{\textbf{0.932}} & \textcolor{red}{\textbf{0.017}} & \textcolor{red}{\textbf{0.944}} & \textcolor{red}{\textbf{0.943}} & \textcolor{red}{\textbf{0.930}} & 0.912 & \textcolor{red}{\textbf{0.023}} & \textcolor{red}{\textbf{0.973}} & \textcolor{blue}{0.952} & \textcolor{red}{\textbf{0.946}} & \textcolor{red}{\textbf{0.948}} & \textcolor{red}{\textbf{0.019}} & \textcolor{red}{\textbf{0.950}} & \textcolor{red}{\textbf{0.917}} & \textcolor{red}{\textbf{0.907}} & \textcolor{red}{\textbf{0.927}} & \textcolor{red}{\textbf{0.033}} & \textcolor{red}{\textbf{0.945}} & \textcolor{red}{\textbf{0.932}} & \textcolor{red}{\textbf{0.909}} & \textcolor{blue}{0.909} & \textcolor{red}{\textbf{0.027}}\\
    \bottomrule
    \hline
  \end{tabular}}
\end{table*}

\subsection{Complexity Analysis}
We present the number of parameters (Params), floating point operations (FLOPs), and frames per second (FPS) of RGB-T SOD models and each component of the proposed \proposed in Tables \ref{tab:RGBTcomparison}, \ref{tab:Ablation}, and \ref{tab:input_size}. 
FPS testing was performed on an NVIDIA RTX 4060 laptop GPU to simulate real-world deployment on edge devices and consumer-grade terminals.
Overall, \proposed is a heavyweight model with a relatively slow inference speed compared to the models in Table \ref{tab:RGBTcomparison}, achieving 17.1 FPS on an NVIDIA RTX4060 Laptop GPU and peaking at 1,114MiB of memory usage.
Compared to the wavelet transform-based WaveNet \cite{10127616}, it has better performance, is 3$\times$ faster in inference, 
and significantly reduces peak memory consumption by 3,711MiB,
despite having 68.9M more Params and 75.5G more FLOPs. 
When compared to the purely Transformer-based UniTR \cite{UniTR}, \proposed has 3.3M more Params, 4.5G more FLOPs, and only half the inference speed; however, it supports a higher input resolution of 384$^2$, whereas UniTR is limited to 224$^2$ and has a peak memory of 1,963MiB.
Specifically, most of \proposed's computational cost comes from the backbone network, while our two proposed components, namely MPA and FEB, contribute minimally. The MPA increases only 5.4M Params and 1.6G FLOPs compared to element-wise addition, whereas the FEB just requires 1.0M Params and 3.9G FLOPs. Additionally, the complexity of FRCAB is nearly identical to RCAB, with a 1.5M increase in Params overall.

\subsection{Limitations and Future Works}
Despite achieving outstanding performance, our \proposed still struggles with some complex scenarios.
\begin{figure}[!htb]
	\centering \includegraphics[width=0.47\textwidth]{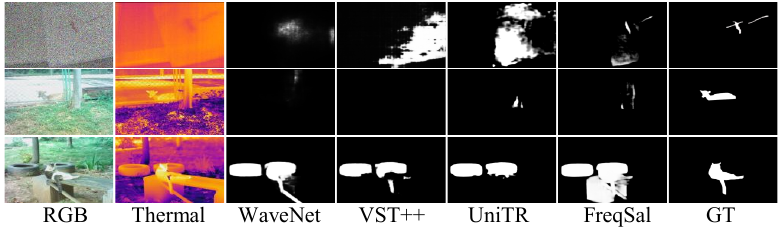}
	\caption{Failure cases of the proposed \proposed. GT denotes ground-truth.}
\label{fig:fail}
\end{figure}
As shown in Fig. \ref{fig:fail}, \proposed erroneously identifies salient objects when the background exhibits more discriminative features than foreground objects, such as higher regional contrast and greater temperature change. 
In contrast, due to the existence of the category concept, it is easier for the human attention mechanism to focus on central objects or living organisms while ignoring cluttered backgrounds.
To mimic this mechanism, future improvements could involve adding prompts such as text and audio, or introducing large models to get more discriminating information about salient objects.

Furthermore, while FFT-based operations demonstrate remarkable potential, their integration with spatial-domain operations remains an open challenge due to compatibility issues. 
The combination of frequency- and spatial-domain techniques improves the model's ability to capture irregular object shapes, highlighting strong potential for future research. This direction can be further advanced by designing normalization layers and optimizers tailored to frequency-domain features as well as by establishing a comprehensive theoretical framework.

\section{Conclusion}
In this paper, we proposed \proposed to explore the potential of 
FFT in solving 
RGB-T SOD.
Specifically, to fully bridge the complementary information and fuse the RGB and thermal modalities, we designed the 
MPA, which embeds channel FFT into spatial Fourier components, achieving deep-level and multi-dimensional representation enhancement. 
Compared to conventional 
Transformer-based operations, MPA can be easily applied to high-resolution bimodal feature fusion. 
In the decoding stage, we proposed the 
FEB and FRCAB
to achieve high-resolution saliency maps. 
Based on the analysis of bimodal frequency decomposition, we presented FEB to clarify accurate object edges from the background and guide the integration of decoder features. We then deployed FRCAB in each decoder layer to focus on high-frequency information and obtain global dependencies in the channel dimension of the frequency domain. 
In the training stage, we proposed the 
CFL to assist FEB in solving hard frequencies and enhance edge representations within the bimodal feature spaces. 
This frequency-domain learning synergized with spatial-domain learning to realize high-quality saliency maps.
Extensive experiments demonstrate that \proposed not only surpasses twenty-nine existing bimodal models on the ten publicly available benchmark datasets but more importantly, reveals the under-explored potential of Fourier-domain representations, potentially paving the way for paradigm evolution in the SOD tasks.

\bibliographystyle{IEEEtran}
\bibliography{ref}

\begin{IEEEbiography}[{\includegraphics[width=1in,height=1.25in,clip,keepaspectratio]{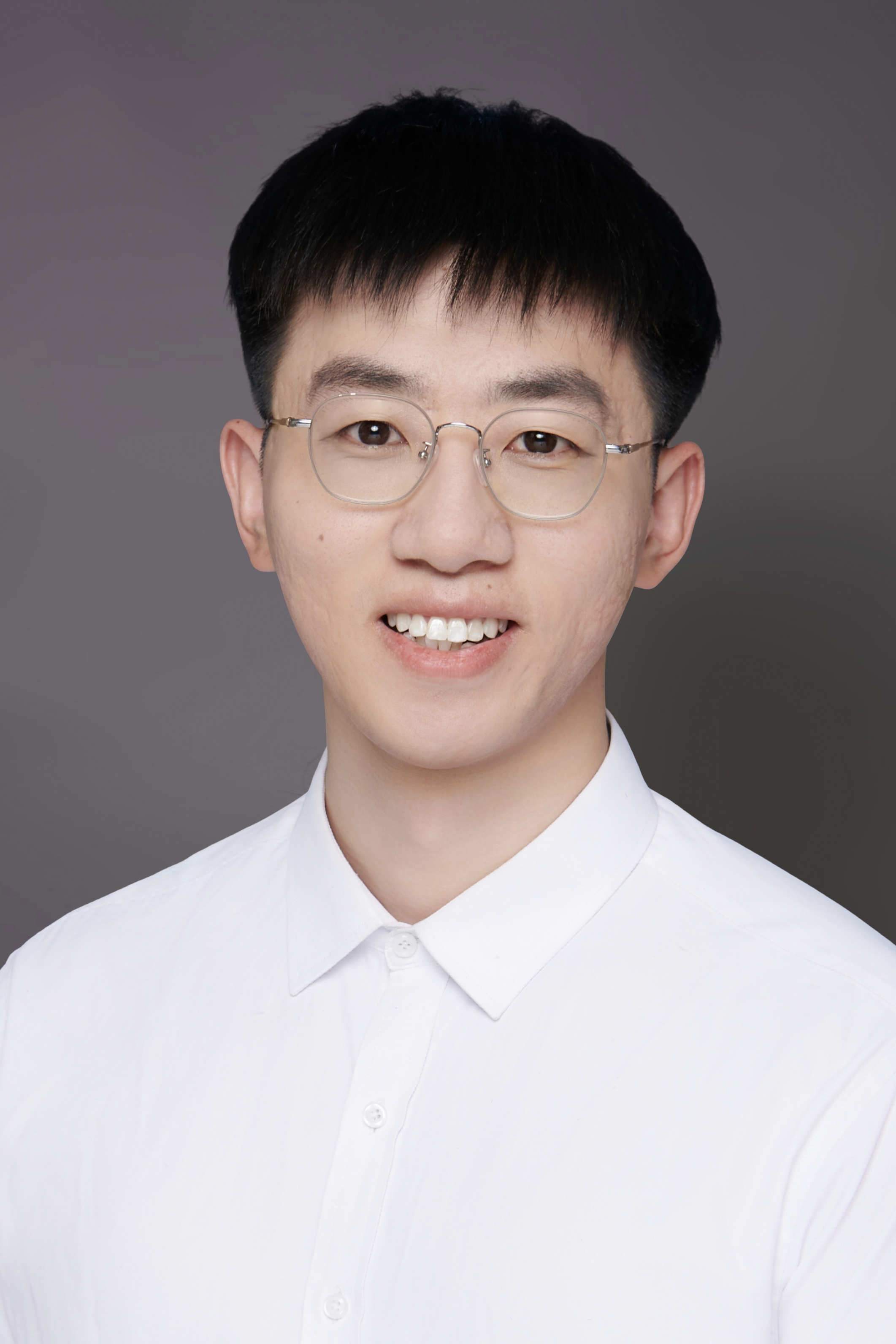}}]{Pengfei Lyu} received M.S. degree in the College of Information Science and Engineering from Northeastern University. He is currently pursuing the Ph.D. degree at the Faculty of Robot Science and Engineering, Northeastern University, Shenyang, China, and simultaneously serves as a joint Ph.D. student at the College of Computing and Data Science, Nanyang Technological University, Singapore. His current research interests include machine learning and its applications in multi-modal nature/medical image processing.
\end{IEEEbiography}

\vspace{-10mm}
\begin{IEEEbiography}
[{\includegraphics[width=1in,height=1.25in,clip,keepaspectratio]{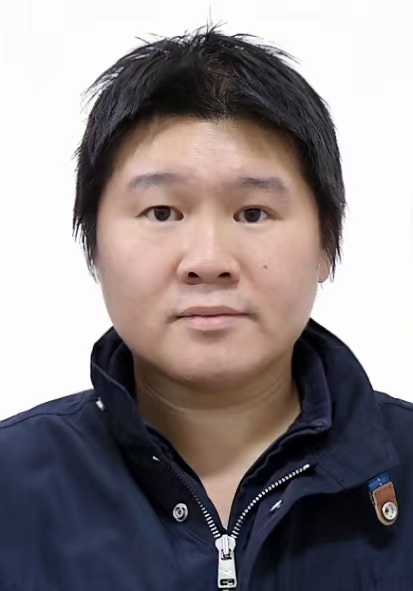}}]{Xiaosheng Yu} received the Ph.D. degree from Northeastern University, Shenyang, China, in 2014. 
He is an associate professor at the Faculty of Robot Science and Engineering, Northeastern University, Shenyang, China. His research interests include medical image processing, salient object detection, and abnormal event detection.
\end{IEEEbiography}

\vspace{-10mm}
\begin{IEEEbiography}
[{\includegraphics[width=1in,height=1.25in,clip,keepaspectratio]{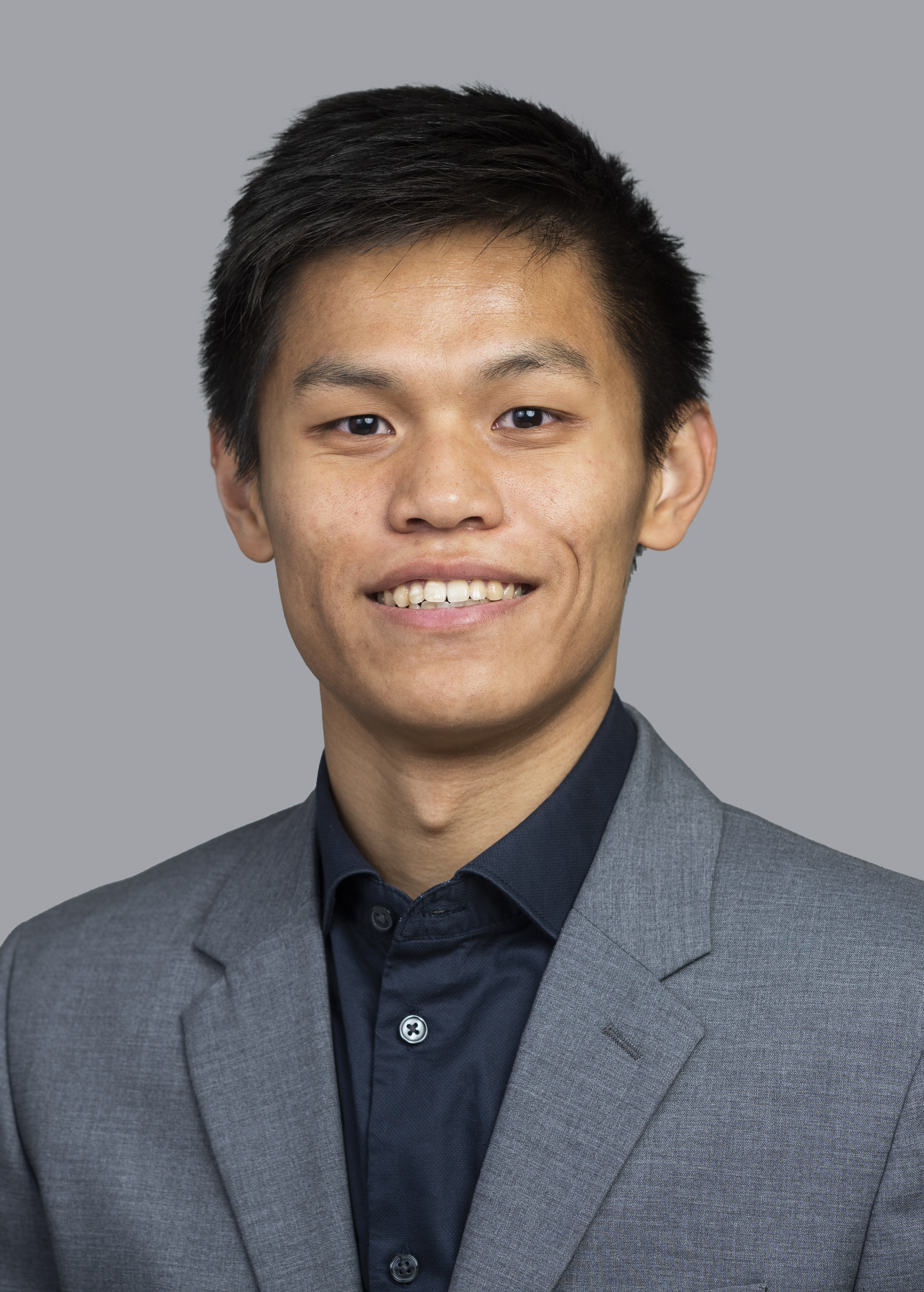}}]
{Pak-Hei Yeung} received the B.Eng. degree in Medical Engineering from the University of Hong Kong in 2018, and the DPhil degree in Engineering Science from University of Oxford in 2023. He is currently a Presidential Postdoctoral Fellow at the College of Computing and Data Science, Nanyang Technological University, Singapore. His current research interests include medical image analysis and computer vision.
\end{IEEEbiography}

\vspace{-10mm}
\begin{IEEEbiography}
[{\includegraphics[width=1in,height=1.25in,clip,keepaspectratio]{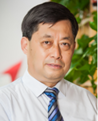}}]{Chengdong Wu} received the B.E. degree in Electronic Automation from Shenyang Jianzhu University, Shenyang, China, and the M.Sc. degree in the Theory and Application of Auto-Control from Tsinghua University, Beijing, China, and received the Ph.D. degree in Automation from Northeastern University, Shenyang, China. 
He is currently a Professor with the Faculty of Robot Science and Engineering, Northeastern University, China. His research interests include image processing, machine vision, deep learning, and intelligent robot systems.
\end{IEEEbiography}

\vspace{-10mm}
\begin{IEEEbiography}
[{\includegraphics[width=1in,height=1.25in,clip,keepaspectratio]{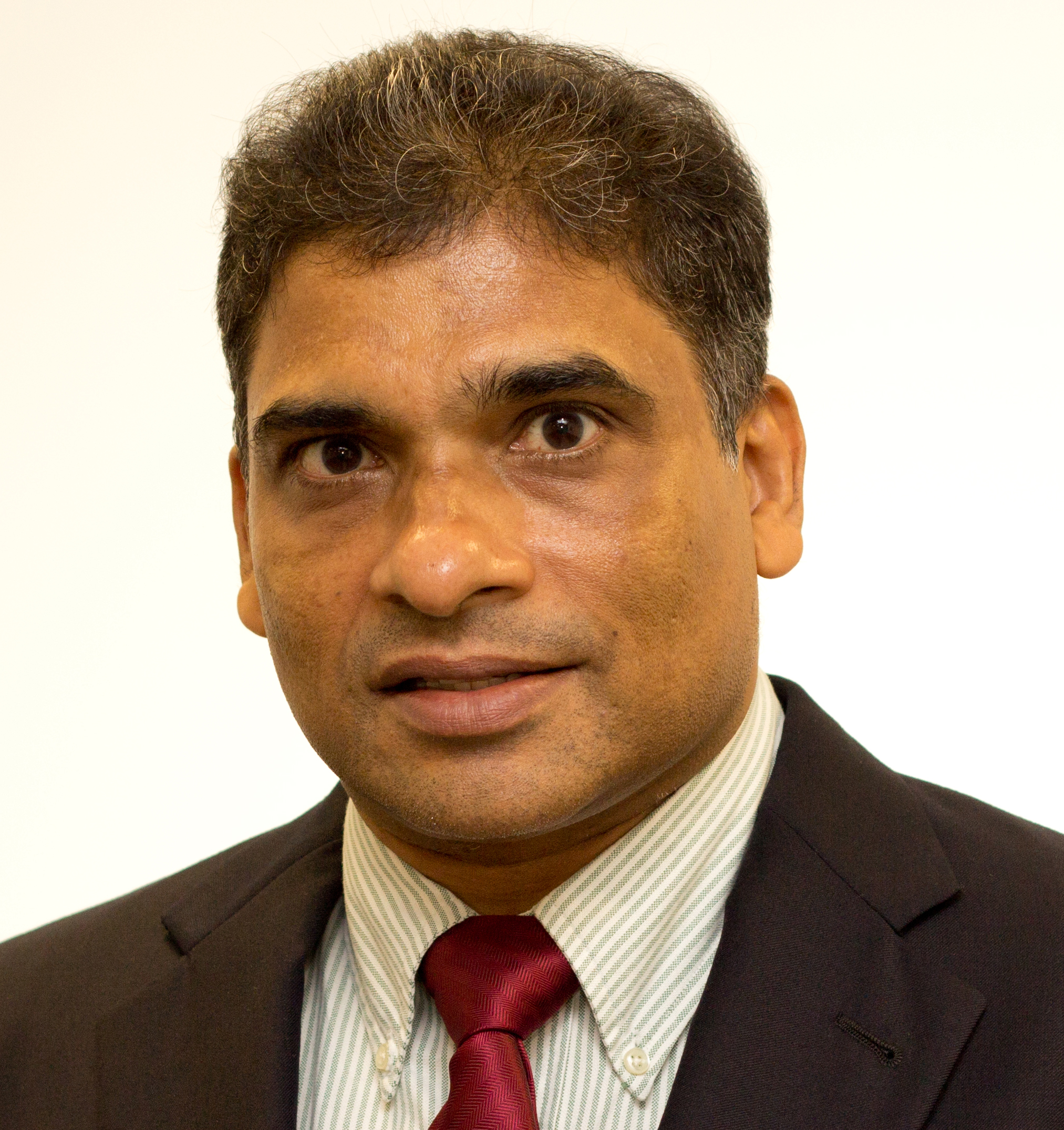}}]{Jagath C. Rajapakse} (Fellow. IEEE) received the BSc (Eng) degree with first-class honors from the University of Moratuwa (UM), Sri Lanka, and the MS and PhD degrees in electrical and computer engineering from the University of Buffalo (UB), USA. 
He was a visiting professor to the Department of Biological Engineering at the Massachusetts Institute of Technology (MIT), USA. He was a visiting fellow at the National Institute of Mental Health and visiting scientist at the Max-Planck-Institute of Brain and Cognitive Sciences, Germany. He is currently a professor of Computer Science and Engineering, Nanyang Technological University (NTU), Singapore. His current research focus on developing techniques and tools for diagnosis and treatment of brain diseases. In these areas, he has published over 300 peer-reviewed research articles in high-impact journals and conferences. 

He serves as Editor for Engineering Applications in Artificial Intelligence and served as Associate Editor for IEEE Transactions on Medical Imaging, IEEE Transactions on Neural Networks and Learning Systems, and IEEE Transactions on Computational Biology and Bioinformatics. He was a Fulbright Scholar and appointed IEEE Fellow in 2012 in recognition of his contributions to brain image analysis.
\end{IEEEbiography}

\newpage

\section*{Appendix}
\label{sec:append}

\subsection{Summary of Transformer-based Accuracy-targeted Model Structures}
\begin{figure}[!htb]
	\centering \includegraphics[width=0.47\textwidth]{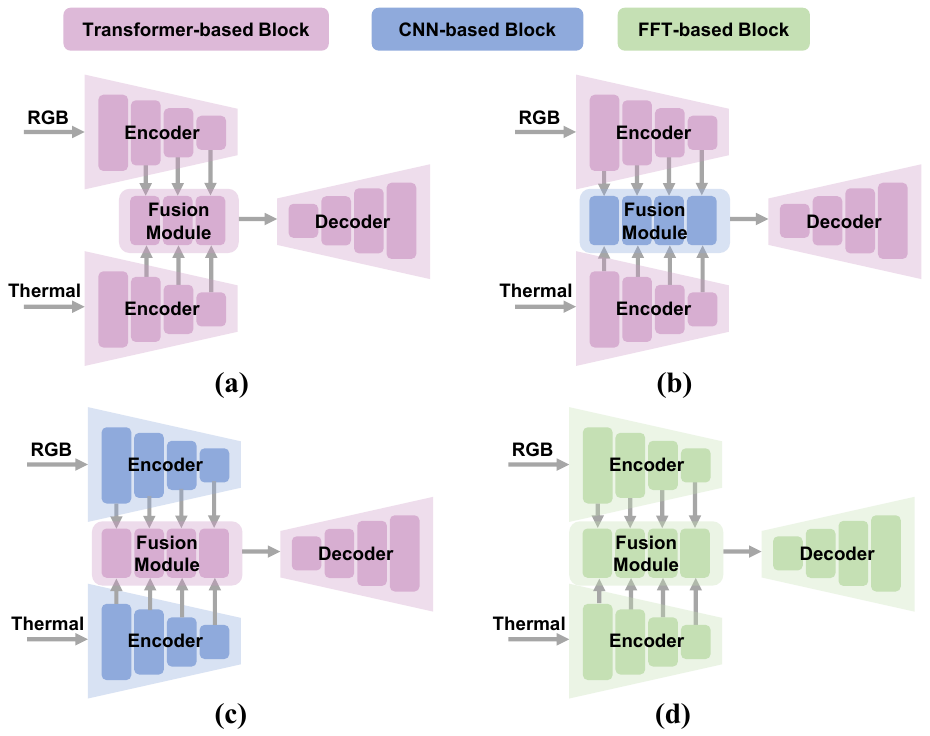}
	\caption{Simplified structure diagrams of existing Transformer-based accuracy-targeted RGB-T SOD models. (a) represents models that ignore high-resolution bimodal features, including VST \cite{VST}, PRLNet \cite{zhou2023position}, VST++ \cite{liu2024vst++}, UniTR \cite{UniTR}, and SACNet \cite{SACNet}. (b) presents models that employ CNN-based fusion modules for bimodal feature interaction, including IFFNet \cite{10015881}, SwinNet \cite{9611276}, CGMDRNet \cite{chen2022cgmdrnet}, HRTNet \cite{9869666}, and CACFNet \cite{jin2024cafcnet}. (c) shows models that employ CNN-based pretrained encoders to extract initial bimodal features, including CGMDRNet \cite{chen2022cgmdrnet}, CACFNet \cite{jin2024cafcnet}, and CAVER \cite{10015667}. Unlike these current models, all components of our FreqSal are designed using FFT (d), allowing it to capture global relationships across all stages.}
\label{fig:TF_structure}
\end{figure}

We summarized the simplified architectures of existing Transformer-based accuracy-targeted RGB-T SOD models in Fig. \ref{fig:TF_structure}. 
To reduce memory usage when training under limited computational resources, these models typically adopt three compromise strategies: (1)
VST \cite{VST}, PRLNet \cite{zhou2023position}, VST++ \cite{liu2024vst++}, UniTR \cite{UniTR}, and SACNet \cite{SACNet} ignore high-resolution
bimodal features; (2) IFFNet \cite{10015881}, SwinNet \cite{9611276}, CGMDRNet \cite{chen2022cgmdrnet}, HRTNet \cite{9869666}, and CACFNet \cite{jin2024cafcnet} employ CNN-based fusion modules; and (3) CGMDRNet \cite{chen2022cgmdrnet}, CACFNet \cite{jin2024cafcnet}, and CAVER \cite{10015667} employ CNN-based pretrained encoders.
Unlike these Transformer-based models, our FreqSal not only effectively fuses high-resolution bimodal features but also employs FFT-based designs to capture global relationships across all stages.

\subsection{Additional Ablation Study}
\emph{1) Effectiveness of activation function in MPA (Sec. \ref{sec:MPA}):}
To validate the effectiveness of StarReLU \cite{yu2022metaformer} as the activation function in the proposed MPA, we compared it with two standard activations, namely ReLU and GeLU, labeled ``StarReLU $\rightarrow$ ReLU'' and ``StarReLU $\rightarrow$ GeLU'' in Table \ref{tab:activation}. The variants with ReLU and GeLU performed worse across all four datasets, confirming StarReLU's superiority.
Compared with ReLU and GeLU, StarReLU produces smoother frequency-domain responses and more stable gradient behavior, as its squaring operation corresponds to the square of the amplitude in the Fourier domain, thereby preventing abrupt truncation of high-frequency components and preserving phase information that mainly resides in these frequencies. 
Moreover, the inclusion of a bias term allows weak activations in negative regions, further supporting phase preservation. 
In addition, StarReLU grows linearly in its non-zero regions, effectively avoiding gradient explosion---an essential property for stable inverse Fourier transforms and gradient backpropagation in the frequency domain. 
Therefore, StarReLU proves more suitable for the proposed MPA, which involves successive Fourier transforms.

\begin{table}[!htp]
  \centering
  \fontsize{8}{10}\selectfont
  \renewcommand{\arraystretch}{1}
  \renewcommand{\tabcolsep}{0.8mm}
  \scriptsize
  \caption{Ablation Study of Activation Function in Modal-coordinated Perception Attention. The Best Results are Labeled \textcolor{red}{\textbf{Red}}.}
\label{tab:activation}
  \scalebox{1}{
  \begin{tabular}{c|cc|cc|cc|cc}
  \hline\toprule
    \multirow{2}{*}{\centering Settings}  & \multicolumn{2}{c|}{\centering VT821} & \multicolumn{2}{c|}{\centering VT1000} & \multicolumn{2}{c|}{\centering VT5000} &\multicolumn{2}{c}{\centering VI-RGBT1500}\\
   \multicolumn{1}{c|}{} & $F_\beta\uparrow$ & $\mathcal{M}\downarrow$ & $F_\beta\uparrow$ & $\mathcal{M}\downarrow$ & $F_\beta\uparrow$ & $\mathcal{M}\downarrow$ & $F_\beta\uparrow$ & $\mathcal{M}\downarrow$ \\
\midrule
    StarReLU $\rightarrow$ ReLU & 0.868 & 0.024 & 0.897 & 0.015 & 0.879 & 0.024 & 0.886 & 0.027\\
    StarReLU $\rightarrow$ GeLU & 0.863 & \textcolor{red}{\textbf{0.023}} & 0.893 & \textcolor{red}{\textbf{0.014}} & 0.881 & 0.024 & 0.884 & \textcolor{red}{\textbf{0.026}} \\	
\midrule
    StarReLU & \textcolor{red}{\textbf{0.872}} & \textcolor{red}{\textbf{0.023}} & \textcolor{red}{\textbf{0.903}} & \textcolor{red}{\textbf{0.014}} & \textcolor{red}{\textbf{0.884}} & \textcolor{red}{\textbf{0.023}} & \textcolor{red}{\textbf{0.893}}& \textcolor{red}{\textbf{0.026}}\\
    \bottomrule
    \hline
  \end{tabular}}
\end{table}

\emph{2) Effectiveness of the input of FEB (Sec. \ref{sec:FEB}):}
To validate the effectiveness of the input selection strategy for FEB, we compared the original setting (\textit{i.e.,} ``T'' and ``F'' as inputs) with three representative settings, and presented the results in Table \ref{tab:input_FEB}. 
Where ``RGB'', ``T'', and ``F'' denote RGB, thermal, and fusion features, respectively. 
The results show that the best performance is achieved when the fusion and thermal features serve as inputs. 
In contrast, incorporating RGB features leads to a significant drop in $F_{\beta}$ metrics. 
As illustrated in Fig. \ref{fig:visedge}, the phase and high-frequency components of RGB images contain not only edge information but also predominantly cluttered background textures. Since edge and texture frequency signals coexist within the same frequency band, introducing RGB features as FEB inputs dilutes edge signals with background noise and leads to blurred edge features or edge misdetection. In summary, we selected thermal and fusion features as inputs for the FEB to obtain clear edge features.

\begin{table}[!htp]
  \centering
  \fontsize{8}{10}\selectfont
  \renewcommand{\arraystretch}{1}
  \renewcommand{\tabcolsep}{0.8mm}
  \scriptsize
  \caption{Ablation Study of the Inputs of Frequency-decomposed Edge-aware Block. The Best Results are Labeled \textcolor{red}{\textbf{Red}}.}
\label{tab:input_FEB}
  \scalebox{1}{
  \begin{tabular}{c|cc|cc|cc|cc}
  \hline\toprule
   \multirow{2}{*}{\centering Setting} & \multicolumn{2}{c|}{\centering VT821} & \multicolumn{2}{c|}{\centering VT1000} & \multicolumn{2}{c|}{\centering VT5000} & \multicolumn{2}{c}{\centering VI-RGBT1500}\\
   & $F_{\beta}\uparrow$ & $\mathcal{M}\downarrow$ & $F_{\beta}\uparrow$ & $\mathcal{M}\downarrow$ & $F_{\beta}\uparrow$ & $\mathcal{M}\downarrow$ &  $F_{\beta}\uparrow$ & $\mathcal{M}\downarrow$\\
\midrule
    T & 0.870 & 0.024 & 0.899 & 0.015 & 0.878 & 0.024 & 0.892 & 0.027 \\	
    F & 0.865 & 0.025 & 0.892 & 0.016 & 0.876 & 0.024 & 0.890 & 0.027\\		
    RGB + T + F & 0.858 & \textcolor{red}{\textbf{0.023}} & 0.882 & 0.017 & 0.871 & 0.024 & 0.886 & \textcolor{red}{\textbf{0.026}}\\
\midrule   
    T + F & \textcolor{red}{\textbf{0.872}} & \textcolor{red}{\textbf{0.023}} & \textcolor{red}{\textbf{0.903}} & \textcolor{red}{\textbf{0.014}} & \textcolor{red}{\textbf{0.884}} & \textcolor{red}{\textbf{0.023}} & \textcolor{red}{\textbf{0.893}} & \textcolor{red}{\textbf{0.026}} \\	
    \bottomrule
    \hline
  \end{tabular}}
\end{table}

\emph{3) Selection of weight parameters $\left\{ {{\lambda_i}} \right\}_{i = 1}^4$ in Eq. (\ref{eq:total_loss}) (Sec. \ref{sec:bidomain}):}
For the weight parameters $\left\{ {{\lambda_i}} \right\}_{i = 1}^4$ of loss functions in Eq. (\ref{eq:total_loss}), we determined their values through careful testing and set them to 1 in this paper. 
The representative values and corresponding results are presented in Table \ref{tab:total_loss}.
Specifically, $\lambda_1$, $\lambda_2$, $\lambda_3$, and $\lambda_4$ respectively control the importance of the saliency loss ${{\cal L}_S}$, the decoding loss ${{\cal L}_{D}}$, the spatial-domain edge loss ${{\cal L}_{E}}$, and the co-focus loss ${{\cal L}_{CFL}}$ in the total loss ${{\cal L}_{total}}$.
When the values of $\lambda_1$ or $\lambda_2$ deviate from 1, the model performance generally declines to varying degrees. 
For edge feature learning in the bi-domain, a higher spatial-domain loss proportion (\textit{i.e.}, $\lambda_3$ is greater than $\lambda_4$) improves the $\mathcal{M}$ metric, while a higher frequency-domain loss proportion (\textit{i.e.}, $\lambda_4$ is greater than $\lambda_3$) enhances the $F_{\beta}$ metric. 
When all $\left\{ {{\lambda_i}} \right\}_{i = 1}^4$ are set to 1, the model achieves the most balanced performance across all four datasets.

\begin{table}[!htp]
  \centering
  \fontsize{8}{10}\selectfont
  \renewcommand{\arraystretch}{1}
  \renewcommand{\tabcolsep}{0.8mm}
  \scriptsize
  \caption{Selection of Weight Parameters $\left\{ {{\lambda_i}} \right\}_{i = 1}^4$ in Eq. (\ref{eq:total_loss}). The Best Results are Labeled \textcolor{red}{\textbf{Red}}.}
\label{tab:total_loss}
  \scalebox{1}{
  \begin{tabular}{cccc|cc|cc|cc|cc}
  \hline\toprule
    \multicolumn{4}{c|}{\centering Parameters} & \multicolumn{2}{c|}{\centering VT821} & \multicolumn{2}{c|}{\centering VT1000} & \multicolumn{2}{c|}{\centering VT5000} &\multicolumn{2}{c}{\centering VI-RGBT1500}\\
   $\lambda_1$ & $\lambda_2$ & $\lambda_3$ & $\lambda_4$ & $F_\beta\uparrow$ & $\mathcal{M}\downarrow$ & $F_\beta\uparrow$ & $\mathcal{M}\downarrow$ & $F_\beta\uparrow$ & $\mathcal{M}\downarrow$ & $F_\beta\uparrow$ & $\mathcal{M}\downarrow$ \\
\midrule
    0.5 & 1 & 1 & 1 & 0.868 & 0.025 & 0.897 & 0.015& 0.877 & 0.024 & 0.890 & 0.027\\	
    1.5 & 1 & 1 & 1 & 0.870 & 0.024 & 0.905 & 0.015 & 0.882 & 0.024 & 0.893 & \textcolor{red}{\textbf{0.026}}\\		
\midrule
    1 & 0.5 & 1 & 1 & 0.870 & 0.024 & 0.899 & \textcolor{red}{\textbf{0.014}} & 0.882 & \textcolor{red}{\textbf{0.023}} & 0.890 & \textcolor{red}{\textbf{0.026}}\\		
    1 & 1.5 & 1 & 1 & 0.872 & 0.024 & 0.906 & 0.015 & 0.886 & 0.024 & 0.889 & 0.027\\		
\midrule
    1 & 1 & 0.5 & 1 & \textcolor{red}{\textbf{0.876}} & 0.024 & \textcolor{red}{\textbf{0.906}} & 0.015 & 0.886 & 0.024 & 0.895 & 0.027\\	
    1 & 1 & 1.5 & 1 & 0.868 & 0.024 & 0.903 & \textcolor{red}{\textbf{0.014}} & 0.880 & \textcolor{red}{\textbf{0.023}} & 0.889 & \textcolor{red}{\textbf{0.026}}\\		
\midrule
    1 & 1 & 1 & 0.5 & 0.867 & \textcolor{red}{\textbf{0.023}} & 0.905 & 0.015 & 0.878 & 0.024 & 0.889& 0.027\\	
    1 & 1 & 1 & 1.5 &  0.874 & 0.024 & 0.899 & 0.016 &  \textcolor{red}{\textbf{0.887}} & 0.024 & \textcolor{red}{\textbf{0.896}} & 0.028\\
\midrule
    1 & 1 & 1 & 1 &0.872 & \textcolor{red}{\textbf{0.023}} & 0.903 & \textcolor{red}{\textbf{0.014}} & 0.884 & \textcolor{red}{\textbf{0.023}} & 0.893 & \textcolor{red}{\textbf{0.026}}\\
    \bottomrule
    \hline
  \end{tabular}}
\end{table}

\subsection{Additional Discussion}
In the paper, we primarily conducted experiments on FreqSal to evaluate core metrics (\textit{e.g.}, $Em$, $Sm$, $wF_\beta$, $F_\beta$, and $\mathcal{M}$) for the bimodal SOD task, which, while indirectly reflecting a model's use of global context, do not explicitly assess its spatial reasoning ability.

In future work, we plan to develop the FFT-based approach and evaluate it on benchmarks specifically designed for spatial reasoning.
By examining the model's ability to understand spatial relationships, including relative object positions and scene-level structural patterns, we aim to provide direct quantitative evidence supporting the generalizability of FFT-based methods in capturing global dependencies.

\end{document}